\documentclass{article}

\usepackage{microtype}
\usepackage{graphicx}
\usepackage{subcaption}
\usepackage{booktabs} 

\usepackage{hyperref}

\usepackage[accepted]{icml2026}

\usepackage{amsmath}
\usepackage{amssymb}
\usepackage{mathtools}
\usepackage{amsthm}
\usepackage{lineno}
\usepackage{longtable}
\usepackage{booktabs} 
\usepackage{multirow}
\usepackage{colortbl}  
\usepackage{xcolor}    
\usepackage{caption}   
\usepackage{pifont}    
\usepackage{placeins}  
\usepackage{afterpage} 
\usepackage{pdfpages}  
\usepackage{tcolorbox} 
\usepackage{enumitem}  
\usepackage{float}     

\usepackage[capitalize,noabbrev]{cleveref}

\theoremstyle{plain}

\theoremstyle{definition}

\theoremstyle{remark}

\usepackage[textsize=tiny]{todonotes}

\icmltitlerunning{Orchestrating Spatial Semantics via a Zone-Graph Paradigm for Intricate Indoor Scene Generation}

\begin{document}

\twocolumn[
  \icmltitle{Orchestrating Spatial Semantics via a Zone-Graph Paradigm for \\
    Intricate Indoor Scene Generation}



  \icmlsetsymbol{equal}{*}

  \begin{icmlauthorlist}
    \icmlauthor{Meisheng Zhang}{pku,msra}
    \icmlauthor{Shizhao Sun}{msra}
    \icmlauthor{Yang Zhao}{sjtu,msra}
    \icmlauthor{Ziyuan Liu}{pku,msra}
    \icmlauthor{Zhijun Gao}{pku}
    \icmlauthor{Jiang Bian}{msra}
  \end{icmlauthorlist}

  \icmlaffiliation{sjtu}{Shanghai Jiao Tong University, Shanghai, China}
  \icmlaffiliation{pku}{Peking University, Beijing, China}
  \icmlaffiliation{msra}{Microsoft Research Asia, Beijing, China}

  \icmlcorrespondingauthor{Shizhao Sun}{shizsu@microsoft.com}

  \icmlkeywords{3D Scene Generation, Large Language Models, Zone-Graph Planning, Spatial Reasoning}

  \vskip 0.15in
  {\centering
  \includegraphics[width=\textwidth]{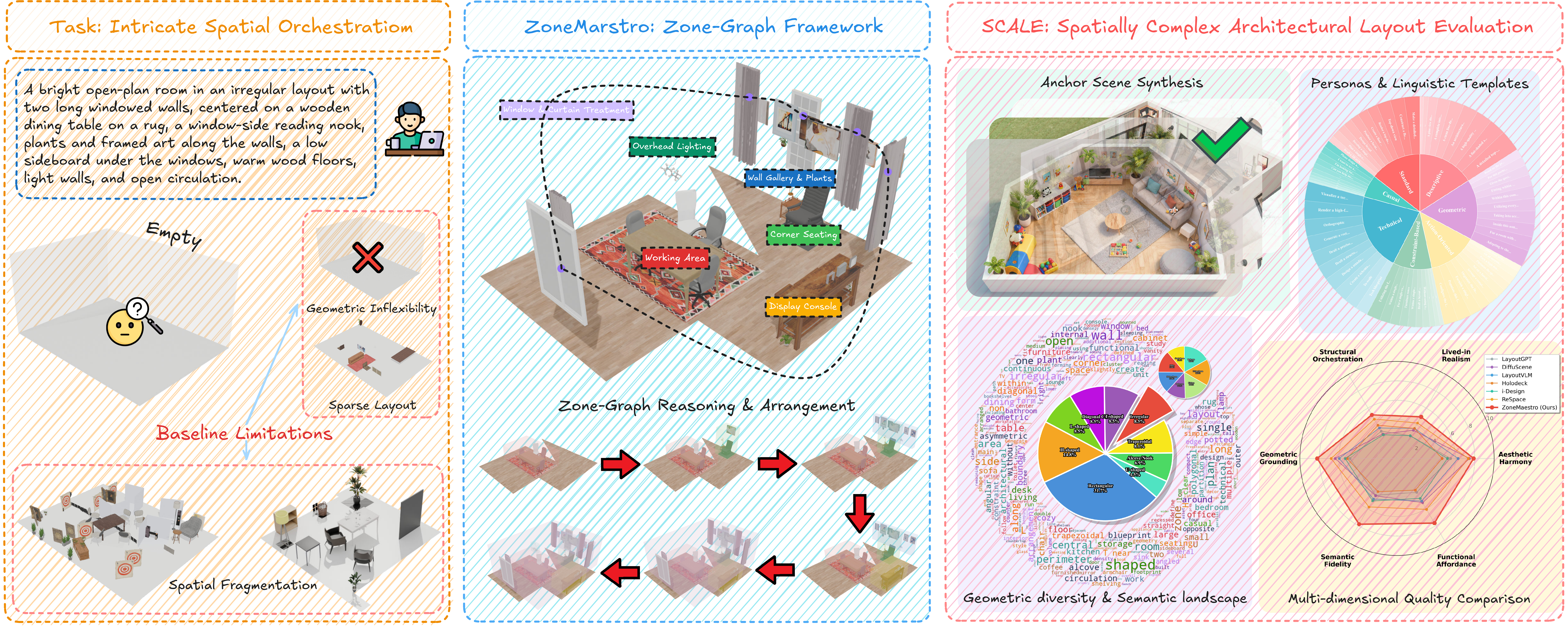}\\[-0.5ex]
  \captionof{figure}{\textbf{Overview of ZoneMaestro.} Left: Existing methods suffer from geometric inflexibility and spatial fragmentation. Middle: Our Zone-Graph framework decomposes the task into compositional reasoning and spatial orchestration. Right: ZoneMaestro achieves superior coherence and density on the SCALE benchmark.}\label{fig:teaser}
  \par}
  \vskip 0.15in
]



\printAffiliationsAndNotice{Work was done during an internship at Microsoft Research Asia.}



\begin{abstract}
Autonomous 3D indoor scene synthesis breaks down in non-convex rooms with tightly coupled spatial constraints. Data-driven generators lack topological priors for long-horizon planning, while iterative agents fragment semantics and become geometrically brittle. We present \textbf{ZoneMaestro}, a unified framework that shifts the paradigm from object-centric synthesis to Zone-Graph Orchestration. By internalizing a novel zone-based logic, ZoneMaestro translates high-level semantic intent into functional zones and topological constraints, enabling robust adaptation to diverse architectural forms. To support this, we construct \textit{Zone-Scene-10K}, a large-scale dataset enriched with explicit Zone-Graph annotations. We further introduce an \textit{Alternating Alignment Strategy} that cycles between reasoning internalization and Zone-Aware Group Relative Policy Optimization (\textit{Z-GRPO}), effectively reconciling the tension between semantic richness and geometric validity without relying on external physics engines. To rigorously evaluate spatial intelligence beyond convex primitives, we formally define the task of \textbf{Intricate Spatial Orchestration} and release SCALE, a stress-test benchmark for irregular indoor scenarios with complex, dense spatial relations. Extensive experiments demonstrate that ZoneMaestro resolves the density-safety dichotomy, significantly outperforming state-of-the-art baselines in both structural coherence and intent adherence.
\end{abstract}

\vspace{-16pt}
\section{Introduction}

Synthesizing 3D indoor environments with semantic and structural fidelity is essential for advancing embodied artificial intelligence, including embodied rearrangement~\cite{wu2023tidybot,gou2024open6dor}, spatial grounding~\cite{chen2024spatialvlm,jatavallabhula2023conceptfusion}, scene graph prediction~\cite{gu2024conceptgraphs}, and long-horizon simulation~\cite{puig2024habitat3,li2024behavior1k}. While language-driven synthesis has mastered simple canonical layouts, its capability diminishes significantly in the regime of Intricate Spatial Orchestration. This domain demands the generation of scenes characterized by dense spatial relationships within irregular, non-convex boundaries. Unlike idealized box-shaped settings, realistic scenarios require managing entangled spatial dependencies rather than mere object quantities, a complexity essential for bridging language instructions with physical spatial realities.
\vspace{-4pt}

Current approaches largely struggle to enforce structural priors in such complex environments. Methods relying on explicit intermediate plans face a grounding gap between abstract relations and metric coordinates, often correcting violations myopically without restoring global structure~\cite{feng2024layoutgpt,layoutvlm2025,wong2025llmtophy3d}. Agentic systems rely on reactive simulation feedback which is often rigid, and multi-step refinement can accumulate deviations that weaken instruction fidelity~\cite{yang2025sceneweaver,yang2024holodeck,celen2024idesign,hu2025marketgen}. Terminal reinforcement learning alignment favors constraint-satisfying shortcuts, locking in early decisions and degrading global coherence~\cite{respace2025,optiscene2025,directlayout2025,metaspatial2025}. Other data-driven generators remain brittle when non-convex boundaries interact with dense inter-object coupling, leading to compounded placement errors~\cite{paschalidou2021atiss,tang2024diffuscene,yang2024physcene,yang2024llplace}.
\vspace{-4pt}

We propose that handling spatial complexity requires a unified \textbf{Zone-Graph Paradigm}. We introduce ZoneMaestro, an LLM-driven framework that reformulates generation through Zone-Graph Reasoning in \cref{fig:teaser}. Unlike methods treating space as a continuous vacuum, ZoneMaestro perceives architectural complexity as a topological graph of functional containers. This \textit{cognitive reconfiguration} enables two capabilities. \textit{Geometric Adaptation} lets zones deform to occupy non-convex recesses. \textit{Semantic Encapsulation} isolates high-density dependencies to prevent the semantic drift that plagues long-horizon autoregression. By internalizing this logic, the model transitions from merely placing objects to curating spatial narratives. To support this, we construct \textit{Zone-Scene-10K}, enriching InternScenes~\cite{zhong2025internscenes} with explicit Zone-Graph reasoning annotations. We further devise an \textit{Alternating Alignment Strategy} cycling between Zone Reasoning Internalization and Zone-Aware Group Relative Policy Optimization (Z-GRPO). By iteratively cycling between intrinsic reward optimization and reasoning consolidation, we prevent the semantic degradation typical of pure RL. This effectively reconciles diverse spatial arrangement with rigorous physical compliance.
\vspace{-4pt}

Formalizing the distinct capabilities required to navigate this regime, we define the task of \textbf{Intricate Spatial Orchestration}. This formulation transcends the simple population of convex hulls, demanding that systems strictly satisfy the topological conflicts between high-density semantic intent and valid geometric execution. However, Existing protocols overlook the critical failures inherent to non-convex regimes \cite{lin2024instructscene,mesatask2025,sceneeval2025}. To drive research beyond current idealized settings, we release \textbf{SCALE}(\textbf{S}patially \textbf{C}omplex \textbf{A}rchitectural \textbf{L}ayout \textbf{E}valuation). By isolating the failure modes of standard baselines, especially their inability to maintain structural coherence within non-convex boundaries, SCALE establishes the first rigorous benchmark for measuring spatial intelligence in realistic, architecturally diverse environments.
\vspace{-4pt}

In summary, our contributions are as follows:
\vspace{-4.5pt}
\begin{itemize}
    \setlength{\itemsep}{2pt}
    \setlength{\parskip}{0pt}
    \item We introduce ZoneMaestro, which reformulates layout synthesis via the \textbf{Zone-Graph Paradigm}. This approach enables semantic encapsulation and geometric adaptation to non-convex boundaries, overcoming the topological myopia of linear baselines.
    \item We devise \textbf{Alternating Spatial Alignment}, cycling between Zone Reasoning Internalization and Zone-Aware GRPO. This reconciles spatial diversity with physical rigor, eliminating geometric noise without semantic degradation.
    \item We formalize the task of Intricate Spatial Orchestration and release \textbf{SCALE}, Spatially Complex Architectural Layout Evaluation. This benchmark isolates failure modes in non-convex topologies, demonstrating our superior structural coherence over existing paradigms.
\end{itemize}
\vspace{-2pt}

\afterpage{%
\begin{figure*}[t]
  \centering
  \vspace{-5pt}  
  \includegraphics[width=0.96\textwidth, trim=0 0 75pt 40pt, clip]{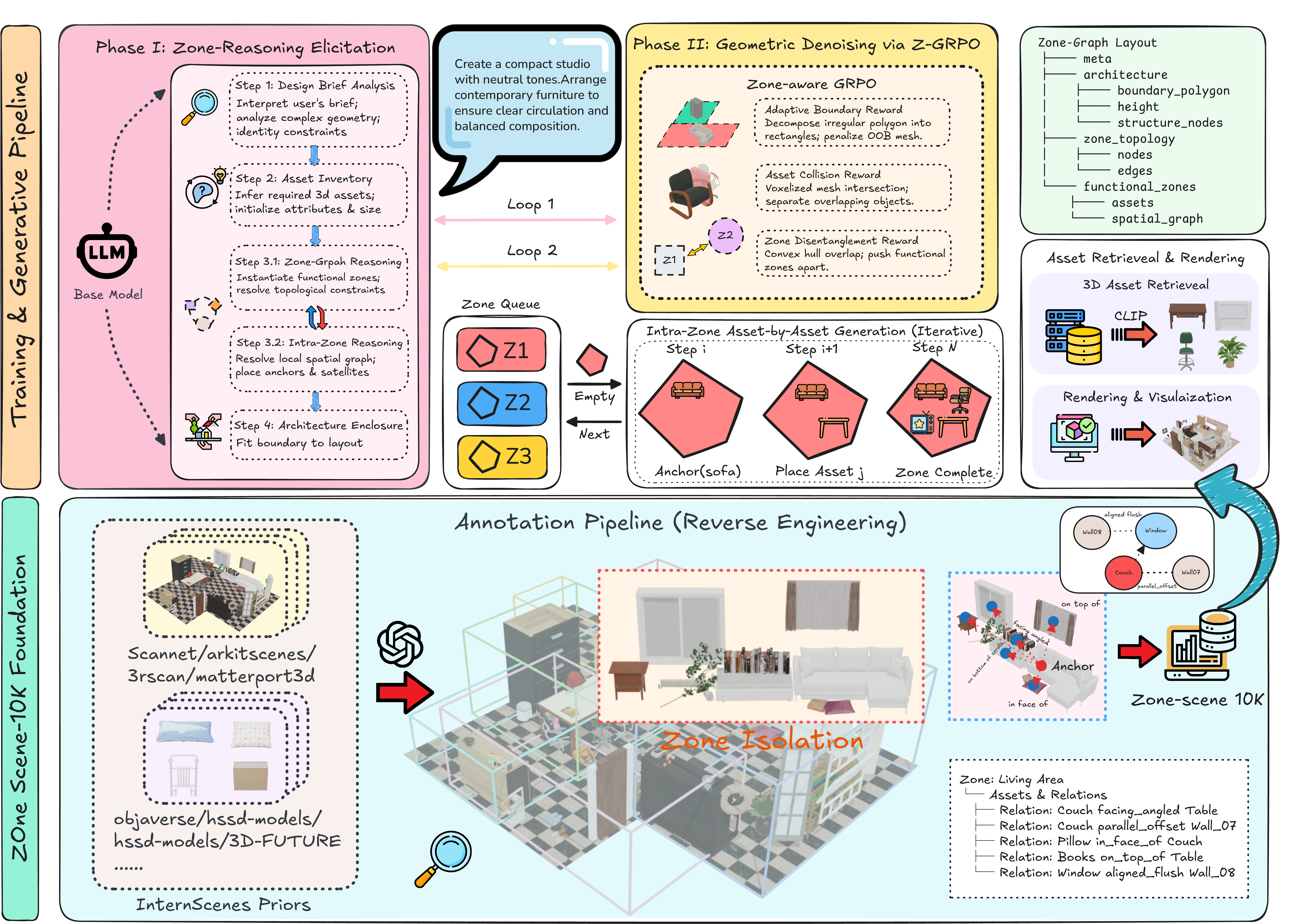}
  \caption{\textbf{The Zone-Graph Paradigm.} Our framework decomposes complex scene generation into four stages: (a) Zone Inventory, (b) Intra-Zone Spatial Graph, (c) Global Topology, and (d) Architecture Derivation.}
  \label{fig:method}
  \vspace{-10pt}  
\end{figure*}
}

\vspace{-5pt}
\section{Related Work}

\textbf{Reasoning-Based Layout Planning.}
Grounding spatial reasoning into metric layouts remains hindered by fragmented workflows. Multi-stage pipelines \cite{respace2025,optiscene2025,directlayout2025} decouple logic from generation, leading to rigid autoregressive ordering \cite{respace2025}, semantic-geometric mismatch \cite{optiscene2025}, or lost stacking relations \cite{directlayout2025}, while reinforcement learning (RL) baselines \cite{metaspatial2025} suffer from exploration inefficiency. Direct generation exposes abstract-to-metric gaps \cite{feng2024layoutgpt} or initialization sensitivity \cite{layoutvlm2025,wong2025llmtophy3d}, and agentic systems \cite{yang2025sceneweaver,yang2024holodeck,celen2024idesign} often fragment semantics across multi-step refinement. Furthermore, other methods exhibit domain rigidity \cite{hu2025marketgen,sun2025roomplanner} or accumulate pixel-level inconsistencies into warped shells \cite{sun20253dgeneralist}. SceneReVis \cite{anonymous2026scenerevis} uses a multi-turn RL formulation to improve generation quality and enable broader applications. In contrast, our Zone-Graph paradigm keeps zone-level structure explicit, linking intent, topology, and metric constraints to ensure global coherence under irregular boundaries.

\vspace{-4pt}
\textbf{Data-Driven 3D Indoor Scene Generation.}
Existing generators struggle to align optimization objectives with scene semantics. Autoregressive frameworks \cite{paschalidou2021atiss} accumulate irreversible placement errors due to missing global planning, while specialized baselines \cite{tang2024diffuscene,yang2024physcene} relying on implicit priors fail strict non-convex boundaries or prioritize local physics over global function. Optimization-driven methods also exhibit gaps: LLplace \cite{yang2024llplace} lacks mechanisms for global functional structure, while RL-based methods like MetaSpatial \cite{metaspatial2025} and ReSpace \cite{respace2025} succumb to reward hacking driven by discriminator inaccuracies or rigid manual rules. Furthermore, pipelines adopting Direct Preference Optimization (DPO) \cite{rafailov2023dpo} such as OptiScene \cite{optiscene2025} and DirectLayout \cite{directlayout2025} tend to overfit canonical patterns, restricting adaptive diversity. In contrast, ZoneMaestro interleaves reasoning internalization with Zone-Aware GRPO to jointly enhance semantic coherence and geometric consistency.

\vspace{-4pt}
\textbf{3D Indoor Scene Datasets \& Evaluation Protocols.}
Data-driven synthesis relies on synthetic repositories, yet 3D-FRONT \cite{fu20213dfront} and Structured3D \cite{zheng2020structured3d} are constrained by sparse arrangements, restricted typologies, or limited scale, while massive aggregations \cite{jia2024sceneverse,zhong2025internscenes} integrate diverse 3D scenes and assets \cite{zheng2020structured3d,xiang2020sapien,dai2017scannet,chang2017matterport3d,baruch2021arkitscenes,wald20193rscan,deitke2023objaverse} but suffer from format heterogeneity and physical defects. This deficit extends to evaluation protocols which remain confined to simplified settings: InstructScene \cite{lin2024instructscene} relies on canonical rectangular layouts, MesaTask \cite{mesatask2025} targets local tabletop rearrangement, and M3DLayout \cite{m3dlayout2025} assesses semantic consistency without probing dense inter-object coupling. Furthermore, methods like PhyScene \cite{yang2024physcene} and SceneEval \cite{sceneeval2025} evaluate semantic reasoning and geometric constraints as separable factors. This motivates SCALE, a benchmark designed to evaluate high-density spatial orchestration under complex boundary profiles and stylized semantic constraints across diverse room typologies.


\vspace{-6pt}
\section{Methodology}

This section presents ZoneMaestro, a unified framework that internalizes the Zone-Graph paradigm shown in \cref{fig:method}. We first model \textbf{Zone-Graph Orchestration} in \cref{sec:problem} and describe Zone-Scene-10K in \cref{sec:dataset}. We then introduce Alternating Spatial Alignment in \cref{sec:phase1}, which interleaves Zone Reasoning Internalization with geometric denoising via Zone-Aware GRPO (Z-GRPO).

\vspace{-12pt}
\subsection{Problem Formulation}
\label{sec:problem}

Central to our approach, we formalize \textbf{Zone-Graph Orchestration} as a conditional generation problem. Given a natural language instruction $\mathcal{X}$, the model outputs a physically valid 3D spatial configuration $\mathcal{S}$. Unlike standard layout tasks that populate a fixed room, our setting requires the model to infer the architectural envelope from the internal zone structure and spatial relations.
\vspace{-3.5pt}

We represent the target scene $\mathcal{S}$ as a compositional tuple $\mathcal{S} = (\mathcal{D}, \mathcal{G}, \mathcal{T}, \mathcal{A})$. The \textit{Zone Inventory} $\mathcal{D}$ defines functional zones and their asset catalogs. The \textit{Intra-Zone Spatial Graph} $\mathcal{G}$ captures local subgraphs where nodes are assets and edges encode spatial constraints. For example, Sofa $\xrightarrow{\text{anchor}}$ Coffee Table describes an anchor relation. The \textit{Global Topology} $\mathcal{T}$ specifies inter zone adjacency. For example, Dining Zone $\xrightarrow{\text{north\_of}}$ Kitchen Zone encodes a relative placement. The \textit{Architecture} $\mathcal{A}$ denotes the derived boundary polygon that encapsulates the assembled topology.
\vspace{-3.5pt}

To model the Zone-Graph paradigm, we factorize the joint probability distribution to reflect this \textit{compositional inference} flow. Local designs determine global topology, which in turn dictates the architecture.
\vspace{-2.5pt}
\begin{equation}
\begin{split}
P(\mathcal{S}, \mathcal{R} | \mathcal{X}) = & P(\mathcal{A} | \mathcal{T}, \mathcal{G}) \cdot P(\mathcal{T} | \mathcal{G}, \mathcal{R}_{\text{topo}}) \\
& \cdot P(\mathcal{G} | \mathcal{D}, \mathcal{R}_{\text{spatial}}) \cdot P(\mathcal{D} | \mathcal{X}, \mathcal{R}_{\text{design}})
\end{split}
\end{equation}
\vspace{-16pt}

\noindent The generation follows a \textit{Design Monologue} $\mathcal{R} = \{\mathcal{R}_{\text{design}}, \mathcal{R}_{\text{spatial}}, \mathcal{R}_{\text{topo}}\}$ that enforces a causal order. The model first determines zone contents $\mathcal{D}$. It then arranges intra-zone assets $\mathcal{G}$, assembles zones into a cohesive whole $\mathcal{T}$, and finally derives the enclosing boundary $\mathcal{A}$. This bottom-up factorization ensures that layouts follow semantic requirements rather than arbitrary placement. A complete input-output example is provided in \cref{sec:appendix_io_example}.

\vspace{-3.5pt}
\subsection{Dataset Construction: Zone-Scene-10K}
\label{sec:dataset}
To support the Zone-Graph Paradigm, we construct Zone-Scene-10K, a large-scale dataset built upon InternScenes~\cite{zhong2025internscenes} and enriched with explicit synthesized Zone-Graph annotations and Reasoning Monologues, as illustrated in \cref{fig:method}. By explicitly grounding latent spatial logic into the data, we provide the necessary supervision for determining functional zones, topological dependencies, and architectural boundaries. The end-to-end construction procedure is summarized in \cref{alg:dataset_pipeline,sec:appendix_instruction}.

\vspace{-13pt}
\paragraph{Zone-Graph Annotation Pipeline.}
Since raw layouts lack functional grouping, we build a Visual-Semantic Decomposition Pipeline that recovers zones and their constraints. Given multi-view renderings, GPT-4o~\cite{openai2024gpt4o} clusters assets into candidate zones and we refine the clusters with split/merge heuristics. For each zone, we render a masked view to extract the intra-zone graph $\mathcal{G}$, and we annotate the global topology $\mathcal{T}$ from zone adjacency and flow. We then generate a Zone-Graph Derivation $\mathcal{R}$ that narrates the same design\,$\rightarrow$\,layout\,$\rightarrow$\,architecture order used by our factorization.

\paragraph{Synthesis of Multi-Granular Design Intents.}
To reflect real user variability, we synthesize instructions at three granularities: \textit{Coarse} descriptions capture overall atmosphere and intent, \textit{Medium} prompts specify category-level furniture lists and composition, and \textit{Fine} prompts state explicit geometric constraints. We further diversify styles and use GPT-4o to generate instructions from rendered views. Full granularity and style definitions are provided in \cref{sec:appendix_instruction}.

\vspace{-28pt}
\paragraph{Reverse-Engineering the Zone-Graph Derivation.}
We synthesize the Zone-Graph Derivation $\mathcal{R}$ by reverse-engineering the ground-truth layout so the trace matches the granularity of $\mathcal{X}$: coarse prompts expand missing inventory and intent, while fine prompts emphasize constraint checking. This produces supervision that ties language, graphs, and geometry at a consistent level of detail.

\vspace{-4.5pt}
\subsection{Alternating Spatial Alignment}
\label{sec:phase1}

We propose Alternating Spatial Alignment, a cyclic optimization that alternates between (i) Reasoning Internalization, which teaches explicit Zone-Graph logic, and (ii) Geometric Denoising, which fixes physical violations via Zone-Aware GRPO. Reciprocal distillation mitigates reward-hacking drift and preserves lived-in realism; see \cref{sec:theory_appendix} for theoretical motivation.

\vspace{-6.5pt}
\paragraph{Zone Reasoning Internalization.}
Unlike flat generation that directly emits object coordinates, we train the model to output a structured reasoning trace that mirrors the design hierarchy. We use coarse-to-fine SFT: the model first predicts the zone inventory and global topology, then instantiates intra-zone graphs into concrete coordinates and orientations. This encourages global planning before local placement.

\vspace{-16pt}
\paragraph{Geometric Denoising via Zone-Aware GRPO.}
We frame geometric refinement as constraint satisfaction solved by Group Relative Policy Optimization. Rather than exploring arbitrary behaviors, Z-GRPO acts as a denoising step that tightens physical compliance with a staged reward. We firstly enforce boundary adherence to ensure containment of assets within the architectural envelope. To handle non-convex room shapes, we implement an adaptive boundary reward that penalizes any footprint that falls outside the decomposed maximal rectangles of the floor plan:
\vspace{-14.5pt}
\begin{equation}
R_{\text{bound}}(Y) = - \lambda_1 \sum_{o_j \in Y} \text{Area}\Big(\text{Box}(o_j) \setminus \bigcup_{m_k \in \mathcal{M}} m_k\Big)
\end{equation}
\vspace{-21pt}

\noindent Second, we maintain separation of functional groups at the intermediate level. We apply a zone disentanglement reward that penalizes the intersection of convex hulls belonging to different zones:
\vspace{-0.5pt}
\begin{equation}
\scalebox{0.935}{$\displaystyle R_{\text{zone}}(Y) = - \lambda_2 \Big( \sum_{z_a \neq z_b} \text{IoU}(\mathcal{H}_{z_a}, \mathcal{H}_{z_b}) + \sum_{z} \text{Area}(\mathcal{H}_z \setminus \mathcal{P}) \Big)$}
\end{equation}
\vspace{-14pt}

Third, we resolve detailed physical intersections at the fine level. The asset collision reward imposes a volumetric penalty on overlapping bounding boxes:
\vspace{-4pt}
\begin{equation}
R_{\text{col}}(Y) = - \lambda_3 \sum_{i \neq j} \text{Vol}(\text{Box}(o_i) \cap \text{Box}(o_j))
\end{equation}
\vspace{-14pt}

\noindent Finally, we drive this targeted relaxation via the composite reward $R = R_{\text{fmt}} + R_{\text{bound}} + R_{\text{zone}} + R_{\text{col}}$:
\begin{equation}
\mathcal{L}_{\text{GRPO}} = \mathbb{E}_{\mathcal{X}} \Big[ \frac{1}{G} \sum_{i} \min \big( r_i \hat{A}_i, \bar{r}_i \hat{A}_i \big) - \beta D_{\text{KL}} \Big]
\end{equation}
\vspace{-14pt}
where $\bar{r}_i = \text{clip}(r_i, 1{-}\epsilon, 1{+}\epsilon)$.

\vspace{4pt}
\paragraph{Zone-Graph Evolution.}
We integrate these two phases into a unified training loop. Each cycle begins with supervised internalization to establish a semantic prior. We then apply geometric alignment to refine physical coordinates. The improved layouts from the reinforcement learning phase are filtered and fed back as training signals for the next round of supervised internalization. This reciprocal distillation improves geometric precision while retaining semantic diversity from expert demonstrations. A theoretical motivation is provided in \cref{sec:theory_appendix}. This loop stabilizes alignment so that optimization does not drift away from the user instruction.

\vspace{-4pt}
\section{Benchmarking Intricate Orchestration}
\label{sec:benchmark}

We now address the critical lack of rigorous evaluation protocols for this domain. Existing benchmarks are saturated with canonical, rectangular layouts that fail to probe a model's ability to orchestrate architectural intricacy~\cite{lin2024instructscene,mesatask2025,m3dlayout2025,yang2024physcene,sceneeval2025}. To bridge this gap, we formalize the task of \textbf{Intricate Spatial Orchestration} and introduce \textbf{SCALE}, a synthetic evaluation suite constructed via a \textit{Visually-Grounded Genesis Pipeline}. The full pipeline figure is in \cref{fig:pipeline,sec:appendix_scale}.

\vspace{-4pt}
\subsection{Task Definition}
\label{sec:task_def}

To ground subsequent evaluation, we define \textbf{Intricate Spatial Orchestration} as generating a full-room layout under two coupled difficulties: (i) geometric irregularity from non-convex boundaries, and (ii) semantic entanglement from high-density constraints with inter-object dependencies. Success requires satisfying both physical feasibility and global functional coherence, rather than optimizing object placement in a canonical convex shell.

\vspace{-4pt}
\subsection{The SCALE Benchmark}
\label{sec:scale_construction}

Building on this task definition, we introduce \textbf{SCALE} for rigorous evaluation. A key challenge is the \textit{grounding gap}: text-only prompt design can be linguistically plausible yet geometrically infeasible. SCALE addresses this by constructing instructions from visually verified layouts, using generative vision systems as a physical plausibility filter. The full pipeline figure is in \cref{fig:pipeline}.
\vspace{-1.5pt}

Concretely, the construction pipeline proceeds through three stages. (1) \textbf{Generation}: we synthesize 22,050 floor plans with the LongCat Image Generator~\cite{longcat2025image}, spanning 9 boundary types including Rectangular, L-shaped, T-shaped, U-shaped, H-shaped, Trapezoidal, a diagonal wall cut, a protruding nook, and other irregular variants as shown in \cref{fig:teaser}. (2) \textbf{Inversion}: we reverse-engineer intent from each image via GPT-4o-mini~\cite{openai2024gpt4o}, producing an initial pool of 22,050 instructions. (3) \textbf{Curation}: we filter by format, deduplicate with CLIP to improve diversity, and stratify sampling, yielding 824 benchmark instances. Construction details appear in \cref{sec:appendix_scale}.

\vspace{-4pt}
\section{Experiments}
\label{sec:experiments}

\afterpage{%
\begin{figure*}[t]
  \centering
  \vspace{-6pt}
  \includegraphics[width=0.98\textwidth]{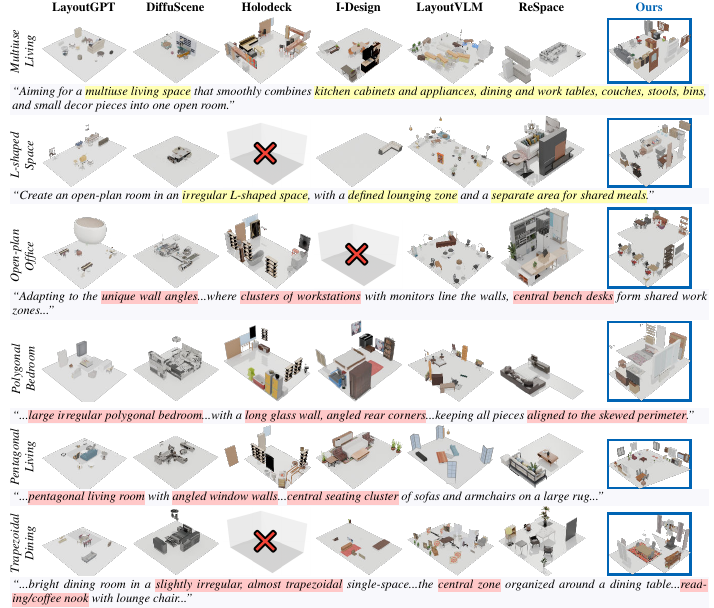}
  \caption{\textbf{Qualitative comparisons.} Rows 1--2: Zone-Scene-10K Test Set. Rows 3--5: SCALE Benchmark featuring complex non-convex geometries and dense asset arrangements. Highlighted instruction phrases often correspond to regions where layouts fail. Agentic frameworks that fail to manage high-density intersections are marked by \textcolor{red}{\ding{55}}. ZoneMaestro follows fine-grained constraints while maintaining valid physical arrangements and alignment with irregular boundaries.}
  \label{fig:qualitative}
  \vspace{-10pt}
\end{figure*}
}

\begin{table*}[t]
\centering
\small
\setlength{\tabcolsep}{2pt}
\caption{Quantitative comparison on the Zone-Scene-10K Test Set and SCALE Benchmark. We evaluate physical validity, semantic quality, and system efficiency. \textbf{Bold} indicates the best performance. GPT-4o-mini scores are reported on a 1--10 scale where higher is better. Human evaluators reports overall preference rankings from our user study where 1 is best and lower is better.}
\label{tab:main_results}
\resizebox{\textwidth}{!}{%
\begin{tabular}{l|ccc|ccccccc|cc}
\toprule
\multirow{2}{*}{\textbf{Method}} & \multicolumn{3}{c|}{\textbf{Physical Validity}} & \multicolumn{7}{c|}{\textbf{Quality Assessment}} & \multicolumn{2}{c}{\textbf{Efficiency}} \\
\cmidrule(lr){2-4} \cmidrule(lr){5-11} \cmidrule(lr){12-13}
 & \textbf{OOB}$\downarrow$ & \textbf{Col.}$\downarrow$ & \textbf{Cnt} & \textbf{Aes.}$\uparrow$ & \textbf{Real.}$\uparrow$ & \textbf{Str.}$\uparrow$ & \textbf{Geo.}$\uparrow$ & \textbf{Sem.}$\uparrow$ & \textbf{Func.}$\uparrow$ & \textbf{Human}$\downarrow$ & \textbf{Succ.}$\uparrow$ & \textbf{Calls}$\downarrow$ \\
\midrule
\multicolumn{13}{c}{\textit{\textbf{Zone-Scene-10K Test Set}}} \\
\midrule
DiffuScene~\cite{tang2024diffuscene} & 0.33 & 0.68 & 9.88 & 6.94 & 3.21 & 2.55 & 6.19 & 3.74 & 4.89 & 4.88 & 100\% & -- \\
ReSpace~\cite{respace2025} & 0.15 & 0.10 & 4.36 & 7.23 & 3.71 & 4.74 & 7.80 & 7.63 & 6.99 & 4.36 & 99.9\% & 5.3 \\
LayoutGPT~\cite{feng2024layoutgpt} & 1.00 & 0.18 & 7.44 & 5.24 & 2.48 & 3.38 & 6.34 & 6.05 & 5.39 & 6.15 & 100\% & 1 \\
LayoutVLM~\cite{layoutvlm2025} & 0.19 & 0.05 & 17.40 & 6.89 & 3.85 & 3.60 & 6.78 & 6.31 & 6.02 & 3.12 & 91.5\% & 3 \\
Holodeck~\cite{yang2024holodeck} & 0.49 & 0.06 & 23.78 & 6.21 & 5.09 & 5.28 & 6.48 & 6.11 & 5.46 & 2.83 & 85.8\% & 6.0 \\
i-Design~\cite{celen2024idesign} & 2.80 & 2.34 & 15.30 & 6.00 & 3.21 & 3.50 & 5.62 & 5.04 & 5.11 & 5.09 & 75.6\% & 9.3 \\
\textbf{ZoneMaestro} & \textbf{0.05} & \textbf{0.03} & 15.51 & \textbf{8.21} & \textbf{4.89} & \textbf{5.61} & \textbf{8.33} & \textbf{8.04} & \textbf{8.14} & \textbf{1.57} & \textbf{99.5\%} & \textbf{1} \\
\midrule
\multicolumn{13}{c}{\textit{\textbf{SCALE Benchmark}}} \\
\midrule
DiffuScene~\cite{tang2024diffuscene} & 0.22 & 0.74 & 10.20 & 7.45 & 3.36 & 2.92 & 6.37 & 4.43 & 5.27 & 5.47 & 100\% & -- \\
ReSpace~\cite{respace2025} & 0.39 & 0.66 & 9.48 & 7.34 & 4.22 & 4.68 & 6.79 & 5.78 & 6.06 & 4.62 & 100\% & 10.7 \\
LayoutGPT~\cite{feng2024layoutgpt} & 6.17 & 0.72 & 14.47 & 5.51 & 2.71 & 3.26 & 4.85 & 4.39 & 4.04 & 5.51 & 100\% & 1 \\
LayoutVLM~\cite{layoutvlm2025} & 0.27 & 0.06 & 20.16 & 6.28 & 3.58 & 3.64 & 5.95 & 4.84 & 5.07 & 3.74 & 94.4\% & 3 \\
Holodeck~\cite{yang2024holodeck} & 0.74 & 0.05 & 21.90 & 4.64 & 3.59 & 3.95 & 5.14 & 3.95 & 3.77 & 2.96 & 72.1\% & 8.4 \\
i-Design~\cite{celen2024idesign} & 1.93 & 4.00 & 15.40 & 5.18 & 2.71 & 2.82 & 5.35 & 5.05 & 4.76 & 4.18 & 68.4\% & 9.5 \\
\textbf{ZoneMaestro} & \textbf{0.09} & \textbf{0.04} & 23.35 & \textbf{7.88} & \textbf{4.95} & \textbf{5.19} & \textbf{8.22} & \textbf{7.82} & \textbf{7.69} & \textbf{1.52} & \textbf{98.7\%} & \textbf{1} \\
\bottomrule
\end{tabular}%
}
\vspace{-8pt}
\end{table*}

\vspace{-2pt}
\subsection{Experimental Setup}
\label{sec:setup}

\vspace{-3pt}
\paragraph{Datasets.}
We use \textbf{Zone-Scene-10K} for training and test, where all scenes are sourced from InternScenes\cite{zhong2025internscenes}. User design instructions and Zone-Graph chains are generated following \cref{sec:dataset}. Each SFT example contains a user design instruction and a reasoning chain with a Zone-Graph layout. We use $N=8{,}500$ for SFT training, $N=500$ for validation, and $N=1{,}000$ for test, with training and validation plus test sampled from disjoint InternScenes sources with uniform room type coverage. For Z-GRPO, we sample $N=5{,}120$ examples from the SFT training set, with $N=2{,}560$ per cycle. We also evaluate on the \textbf{SCALE Benchmark} with $N=824$ instances, and both test protocols provide only user design instructions.

\vspace{-6pt}
\paragraph{Implementation Details.}
We use Qwen3-8B~\cite{qwen3} and train on 8 NVIDIA A100 GPUs. Training follows two cycles of Alternating Alignment. For internalization, we run SFT for 2 epochs per cycle with global batch size 8 and 8 gradient accumulation steps. For geometric alignment, we run Z-GRPO with batch size 32 for 40 optimization steps. We use AdamW with learning rates $1e^{-5}$ for SFT and $5e^{-6}$ for Z-GRPO, with KL coefficient $\beta=0.04$.

\vspace{-6pt}
\paragraph{Baselines.}
We compare ZoneMaestro against: (1) \textbf{Data-Driven Methods} including DiffuScene~\cite{tang2024diffuscene} and ReSpace~\cite{respace2025}, using their official pre-trained checkpoints; and (2) \textbf{Agentic Frameworks} including LayoutGPT~\cite{feng2024layoutgpt}, LayoutVLM~\cite{layoutvlm2025}, Holodeck~\cite{yang2024holodeck}, and i-Design~\cite{celen2024idesign}, powered by GPT-4o~\cite{openai2024gpt4o}. We exclude OptiScene~\cite{optiscene2025} and DirectLayout~\cite{directlayout2025} due to unavailable code and datasets, MetaSpatial~\cite{metaspatial2025} for requiring auxiliary scene information, and SceneWeaver~\cite{yang2025sceneweaver} given its substantial inference-time cost reported, making large-scale evaluation impractical.

\vspace{-10pt}
\paragraph{Evaluation Metrics.}
We report three metric categories for systematic comparison. For \textbf{physical validity}, we measure Out-of-Bounds volume (OOB) for asset protrusion beyond the room boundary, collision volume (Col) for volumetric intersection between assets, and asset count (Cnt) for scene density. For \textbf{semantic quality}, a GPT-4o-mini judge rates layouts on a 1--10 scale across six dimensions: \textit{Aesthetics} (Aes) for visual appeal, \textit{Realism} (Real) for lived-in plausibility, \textit{Structure} (Str) for zone hierarchy and functional organization, \textit{Geometry} (Geo) for adaptation to non-convex boundaries, \textit{Semantics} (Sem) for instruction adherence, and \textit{Functionality} (Func) for ergonomic usability. We additionally report \textbf{Human} overall preference ranking (Human) from our user study to complement automated judgement and capture aspects that may be missed by GPT-based scoring. For \textbf{efficiency}, we report generation success rate (Succ) and LLM inference calls per scene.

\vspace{-4pt}
\subsection{Main Results}
\label{sec:results}

We evaluate ZoneMaestro against leading baselines on both the held-out Test Set from Zone-Scene-10K and the SCALE Benchmark. Quantitative results are summarized in \cref{tab:main_results}.

\vspace{-7.5pt}
\paragraph{Performance on Zone-Scene-10K.}
The standard test set reveals a density-validity tension across all baselines. ReSpace keeps violations low with OOB/collision 0.15/0.10, but generates only 4.36 assets per scene and scores 3.71 on realism, producing sparse and sterile rooms. DiffuScene reaches 9.88 assets but suffers the worst collision volume at 0.68. Among agentic methods, i-Design collapses physically with OOB 2.80 and collision 2.34. LayoutGPT also breaks boundaries with OOB 1.00 and has the lowest aesthetics at 5.24, while LayoutVLM shows weak zoning logic with structure 3.60. Holodeck reaches high density at 23.78 assets but still sacrifices validity and tops out at 5.46 on functionality. ZoneMaestro is the only one that stays both dense and clean, generating 15.51 assets with OOB/collision 0.05/0.03, while ranking best on aesthetics, structure, and functionality.

\vspace{-13pt}
\paragraph{Performance on the SCALE Benchmark.}
SCALE is substantially harder due to non-convex shells and dense constraints. Methods stable on standard rooms degrade quickly: ReSpace more than doubles its boundary errors, from 0.15 to 0.39; LayoutGPT fails most dramatically with OOB surging from 1.00 to 6.17; i-Design loses control of intersections with collision volume reaching 4.00. ZoneMaestro remains stable, keeping OOB at 0.09 and collision at 0.04 while producing the highest asset density at 23.35. The quality gap is also structural rather than purely geometric. DiffuScene drops to 2.92 on structure, and LayoutVLM falls to 3.58 on realism, revealing that complex boundaries amplify semantic fragmentation. Even Holodeck, which keeps collision low, increases its boundary errors to 0.74 and still fails to deliver usable dense arrangements.

\vspace{-13pt}
\paragraph{Efficiency vs. Complexity Analysis.}
Methods relying on test-time adaptation pay a large cost in repeated calls: ReSpace and i-Design require 10.7 and 9.5 inference calls per scene on average. ZoneMaestro uses a single inference pass and achieves a 98.7\% success rate because geometric reasoning is internalized during training.

\begin{table*}[t]
\centering
\small
\setlength{\tabcolsep}{3pt}
\caption{Ablation Study on the SCALE Benchmark. We systematically analyze the contribution of each design choice including Zone-Graph Internalization and the Alternating Alignment strategy. GPT-4o-mini scores are reported on a 1 to 10 scale.}
\label{tab:ablation}
\resizebox{\textwidth}{!}{%
\begin{tabular}{l|ccc|ccccccc}
\toprule
\multirow{2}{*}{\textbf{Variant}} & \multicolumn{3}{c|}{\textbf{Physical Validity}} & \multicolumn{7}{c}{\textbf{GPT-4o-mini Scores (1--10 scale)}} \\
\cmidrule(lr){2-4} \cmidrule(lr){5-11}
 & \textbf{OOB} $\downarrow$ & \textbf{Col.} $\downarrow$ & \textbf{Count} & \textbf{Aes.} $\uparrow$ & \textbf{Real.} $\uparrow$ & \textbf{Struct.} $\uparrow$ & \textbf{Geo.} $\uparrow$ & \textbf{Sem.} $\uparrow$ & \textbf{Func.} $\uparrow$ & \textbf{Avg.} $\uparrow$ \\
\midrule
Base Reasoning SFT w/o Zone-Graph & 0.25 & 0.18 & 38.45 & 7.96 & 3.97 & 6.17 & 7.77 & 7.21 & 7.36 & 6.74 \\
\midrule
Zone-Graph SFT Only & 0.21 & 0.13 & 30.96 & 7.97 & 3.89 & 6.23 & 8.19 & 7.26 & 7.50 & 6.84 \\
Z-GRPO Only w/o Alternating & 0.11 & 0.05 & 21.35 & 7.95 & 4.14 & 6.28 & 8.04 & 7.24 & 7.45 & 6.85 \\
Alternating Alignment One Cycle & 0.14 & 0.05 & 22.30 & 7.90 & 4.05 & 6.15 & 8.00 & 7.10 & 7.30 & 6.75 \\
\midrule
Full Framework ZoneMaestro & 0.09 & 0.04 & 23.35 & 7.88 & 4.95 & 5.19 & 8.22 & 7.82 & 7.69 & 6.96 \\
\bottomrule
\end{tabular}%
}
\vspace{-8pt}
\end{table*}

\vspace{-4pt}
\subsection{User Study}
\label{sec:user_study}

We conducted a user study to quantify perceptual quality in irregular environments. 10 participants ranked seven methods by overall preference on 140 anonymized layouts, with 9 stratified SCALE instances covering all boundary types \cref{sec:scale_construction} and 5 randomly sampled scenarios from the Zone-Scene-10K test set per person. As reported in the Human column of \cref{tab:main_results}, ZoneMaestro ranks first on SCALE with 1.52 and remains best on the Zone-Scene-10K test set with 1.57. Human preference is not identical to GPT trends. On SCALE, ReSpace receives higher GPT realism and semantics at 4.22 and 5.78 than Holodeck at 3.59 and 3.95, yet Holodeck is ranked second by humans at 2.96 while ReSpace drops to 4.62. LayoutGPT ranks last at 5.51. This gap shows ZoneMaestro aligns better with human overall preference while addressing cases where GPT-based scores can be misleading.

\vspace{-4pt}
\subsection{Qualitative Analysis}
\label{sec:qualitative}

We provide visual comparisons across two evaluation settings in \cref{fig:qualitative}: the Zone-Scene-10K Test Set (Rows 1--2) evaluates generalization to diverse real world scenarios, while the SCALE Benchmark (Rows 3--6) targets intricate geometries. From both settings, we highlight three dimensions where ZoneMaestro improves over prior methods. See \cref{sec:appendix_case_gallery} for additional qualitative cases.

\vspace{-13.5pt}
\paragraph{Density-Validity Breakthrough.}
Current paradigms struggle to reconcile asset density with physical validity. In the Open-plan Office scenario optimization-based agents like I-Design fail to converge on massive collision constraints and result in invalid states marked by the cross symbol. LayoutGPT avoids conflicts by generating sparse and disconnected clusters. In the Multiuse Living scene the baselines miss the functional density required by the prompt. ZoneMaestro orchestrates over 20 assets in a single pass without collision. It forms distinct workstation clusters in the office and separates kitchen utilities from the dining zone in the living room. This confirms that the internalized Zone-Graph paradigm effectively buffers the cognitive load of massive arrangements to maintain structural clarity.
\vspace{-10.5pt}
\paragraph{Geometric Intelligence in Non-Convex Spaces.}
Irregular geometries expose the rigidity of heuristic planners. In Row 2 and Row 6 Holodeck fails to navigate the reentrant corners or narrowing widths. Similarly ReSpace and LayoutVLM struggle with the slanted perimeter in the Polygonal Bedroom in Row 4 by placing beds that intersect walls due to axis-aligned biases. ZoneMaestro exhibits precise geometric grounding by aligning large furniture strictly with adjacent wall normals. It utilizes irregular nooks for secondary functions and treats the boundary shape as a guiding constraint rather than an obstacle.
\vspace{-10.5pt}
\paragraph{Emergent Zone Topology.}
ZoneMaestro demonstrates superior topological planning in eccentric spaces beyond obstacle avoidance. In Row 5 baselines like LayoutVLM and ReSpace scatter objects randomly along the walls and rely on alignment heuristics that fail in pentagonal shapes. ZoneMaestro generates a coherent central seating cluster anchored by the rug independent of irregular wall angles. In Row 6 ZoneMaestro successfully distinguishes a reading nook from the primary dining area, while Holodeck fails on the tapered geometry. This behavior validates the Zone-Graph Paradigm, prioritizing functional connectivity over absolute coordinates to preserve human-centric circulation in atypical floor plans.

\vspace{-4pt}
\subsection{Ablation Studies}
\label{sec:ablation}

To validate the contribution of each component in our framework, we conduct ablation studies on the SCALE benchmark. Results are detailed in \cref{tab:ablation}.

\vspace{-10pt}
\paragraph{Internalized Reasoning as Structural Prior.}
We assess the role of Zone-Graph reasoning by training a variant \textit{without Zone-Graph Derivation} that directly outputs layout JSON from instructions. In \cref{tab:ablation}, this variant fails to regulate density and generates 38.45 assets per scene. The inflated count leads to large conflicts, with collision and OOB volumes rising to 0.18 and 0.25. It also lacks coherent functional planning and receives a realism score of 3.97. Zone-Graph SFT Only reduces the count to 30.96 with zone-level allocation and lowers OOB/Col to 0.21/0.13 without RL. This indicates that the zone-structured trace suppresses over-packing and boundary drift, providing a structural prior for constraint satisfaction.

\vspace{-14.0pt}
\paragraph{Counteracting Reward Hacking.}
We then study training stability under reward optimization. Z-GRPO Only without Alternating reduces the asset count to 21.35 and improves collision metrics, but tends to produce cleaner yet less expressive scenes. Alternating Alignment One Cycle recovers density to 22.30, but judge scores remain below Z-GRPO Only. ZoneMaestro runs two full cycles and reaches 23.35 assets, achieving the highest \textbf{Realism (4.95)} and geometric validity. Notably, we observe an inverse correlation between Realism and Structure scores where ablated variants favor rigid grid-aligned layouts with high Structure, while ZoneMaestro optimizes for organic collision-free arrangements to maximize Realism. Detailed analysis of this trade-off is provided in \textbf{Appendix~\ref{app:structure_realism}}.

\vspace{-4pt}
\section{Conclusion}
\vspace{-3pt}

We presented ZoneMaestro, a framework that internalizes Zone-Graph reasoning as a structural prior to preserve topological intent while adhering to non-convex geometric constraints. Our Alternating Alignment strategy effectively resolves the density-safety trade-off, bridging the gap between high-level planning and low-level physical execution. Furthermore, by releasing the SCALE Benchmark, we provide a rigorous testbed to advance 3D layout synthesis beyond canonical convex primitives.


\vspace{-4pt}
\section*{Impact Statement}

This paper presents work whose goal is to advance the field of Machine Learning and 3D scene generation for embodied AI applications. We commit to releasing our code and the Zone-Scene-10K dataset upon acceptance to foster community collaboration and reproducibility. While enabling scalable environment synthesis this technology relies on training data that may reflect specific cultural or geographic architectural norms. There is a risk that the model propagates these biases by overrepresenting Western residential patterns while marginalizing diverse global living styles. We encourage practitioners to curate diverse datasets to mitigate such exclusion. Furthermore automated design tools carry implications for creative employment. We envision ZoneMaestro as an assistive system that enhances human productivity rather than replacing professional expertise. Users should exercise caution to prevent the generation of misleading virtual environments used for deceptive purposes.


\bibliography{custom}
\bibliographystyle{icml2026}

\newpage
\appendix
\onecolumn

\noindent This appendix is organized as follows. \textbf{Appendix~\ref{sec:theory_appendix}} provides theoretical motivation for the Alternating Alignment strategy. \textbf{Appendix~\ref{sec:appendix_dataset}} details the Zone-Scene-10K dataset construction, including data sourcing, instruction synthesis, and statistics. \textbf{Appendix~\ref{sec:appendix_scale}} describes the SCALE benchmark construction pipeline. \textbf{Appendix~\ref{sec:appendix_impl_prompts}} presents implementation details such as the Zone-Graph annotation schema and training hyperparameters. \textbf{Appendix~\ref{sec:appendix_qual}} contains additional qualitative results and an extended case gallery. \textbf{Appendix~\ref{sec:appendix_prompts}} provides the complete prompt collection used throughout the framework. \textbf{Appendix~\ref{sec:appendix_limitations}} discusses limitations and future directions.

\vspace{1em}

\section{Theoretical Motivation for Alternating Alignment}
\label{sec:theory_appendix}

A natural question arises: why adopt an alternating alignment paradigm (Internalization $\rightarrow$ Alignment) rather than jointly optimizing semantic and geometric objectives from the start? We provide intuitive analysis grounded in gradient dynamics and generalization considerations.

\subsection{Gradient Conflict in Joint Optimization}

\paragraph{Observation 1 (Gradient Conflict).}
\textit{The gradients of semantic and geometric objectives tend to exhibit negative cosine similarity during early training, leading to destructive interference.}

Formally, let $g_{\text{sem}} = \nabla_\theta \mathcal{L}_{\text{sem}}$ and $g_{\text{geo}} = \nabla_\theta \mathcal{L}_{\text{geo}}$. We observe that:
\begin{equation}
\cos(g_{\text{sem}}, g_{\text{geo}}) = \frac{\langle g_{\text{sem}}, g_{\text{geo}} \rangle}{\|g_{\text{sem}}\| \|g_{\text{geo}}\|} < 0
\end{equation}
during the initial optimization phase. This occurs because the semantic objective encourages generating rich, diverse layouts (high entropy in asset placement), while the geometric objective penalizes any placement near boundaries or other objects, favoring sparse, conservative solutions. The resulting gradient $g_{\text{joint}} = g_{\text{sem}} + \lambda g_{\text{geo}}$ has reduced magnitude $\|g_{\text{joint}}\| < \|g_{\text{sem}}\|$, slowing convergence and often converging to suboptimal saddle points.

\subsection{Generalization Bounds via Curriculum Learning}

\paragraph{Intuition 2 (Alternating Alignment as Curriculum Learning).}
\textit{Decoupled training can achieve tighter generalization bounds by decomposing the hypothesis class.}

Consider the composite hypothesis class $\mathcal{H} = \mathcal{H}_{\text{sem}} \circ \mathcal{H}_{\text{geo}}$ where semantic reasoning precedes geometric grounding. By the PAC-Bayes framework, the generalization error of a composite learner satisfies:
\begin{equation}
\mathcal{E}_{\text{hier}} \leq \underbrace{\mathcal{E}(\mathcal{H}_{\text{sem}})}_{\text{Internalization}} + \underbrace{\mathcal{E}(\mathcal{H}_{\text{geo}} | \mathcal{H}_{\text{sem}})}_{\text{Alignment}}
\end{equation}
where the conditional complexity $\mathcal{E}(\mathcal{H}_{\text{geo}} | \mathcal{H}_{\text{sem}})$ is significantly smaller than the joint complexity $\mathcal{E}(\mathcal{H})$ because the Alignment step operates on a \textit{semantically anchored} representation rather than raw inputs. Intuitively, once the model has internalized Zone-Graph structure, the geometric refinement becomes a lower-dimensional optimization problem--adjusting coordinates within established functional zones rather than jointly discovering both semantics and geometry.

\subsection{Information-Theoretic Interpretation}
From an information bottleneck perspective, Internalization compresses the instruction $\mathcal{X}$ into a condensed semantic representation $\mathcal{Z} = (\mathcal{D}, \mathcal{G}, \mathcal{T})$ that is maximally informative about the design intent while discarding geometric noise. The Alignment step then maps $\mathcal{Z} \rightarrow \mathcal{A}$ with geometric constraints, operating on a cleaner, lower-entropy input. This decoupled compression-then-refinement mirrors the rate-distortion optimal coding strategy:
\begin{equation}
I(\mathcal{X}; \mathcal{A}) \leq I(\mathcal{X}; \mathcal{Z}) + I(\mathcal{Z}; \mathcal{A} | \mathcal{X})
\end{equation}
Joint training conflates these information channels, forcing the model to simultaneously preserve semantic detail and satisfy geometric hard constraints---objectives that compete for representational capacity.

\section{Zone-Scene-10K Dataset Details}
\label{sec:appendix_dataset}

This appendix provides comprehensive details of the Zone-Scene-10K dataset construction, including data sourcing and curation, instruction synthesis, and dataset statistics.

\subsection{Data Sourcing and Curation}
\label{sec:appendix_sourcing}

We integrate raw scenes from InternScenes and 3D-FRONT to ensure authentic coverage. To target Intricate Spatial Orchestration, we apply a rigorous filtration strategy focusing on \textit{Geometric Complexity} (prioritizing non-convex boundaries and multiple functional areas) and \textit{Asset Density} (discarding sparse scenes), while ensuring balanced \textit{Typology Coverage} across seven room categories to address the scarcity of service spaces.

\subsection{Instruction Synthesis Details}
\label{sec:appendix_instruction}

\paragraph{Zone-Scene-10K Construction Pipeline.}
\cref{alg:dataset_pipeline} summarizes the end-to-end dataset construction procedure.

\begin{algorithm}[t]
\caption{Zone-Scene-10K Construction Pipeline}
\label{alg:dataset_pipeline}
\begin{algorithmic}
   \STATE {\bfseries Input:} raw layouts $\mathbb{L}_{raw}$, VLM $\mathcal{V}$
   \STATE {\bfseries Output:} annotated dataset $\mathcal{D}_{ZS}$
   \STATE \COMMENT{$\textsc{Render}$: multi-view rendering; $\textsc{Group}$/$\textsc{IGraph}$/$\textsc{Topo}$: VLM annotators; $\textsc{Refine}$: split/merge heuristics}
   \STATE \COMMENT{$\textsc{Prompt}$/$\textsc{Derive}$: multi-granular instruction synthesis + reverse-engineered derivation}
   \STATE $\mathcal{D}_{ZS} \leftarrow \emptyset$
   \FOR{each layout $L \in \mathbb{L}_{raw}$}
     \STATE \COMMENT{\textbf{S1: Visual-Semantic Decomposition}}
     \STATE $I \leftarrow \textsc{Render}(L)$
     \STATE $\mathcal{Z} \leftarrow \textsc{Refine}(\textsc{Group}_{\mathcal{V}}(I))$
     \STATE \COMMENT{\textbf{S2: Zone-Graph Extraction}}
     \FOR{each zone $z \in \mathcal{Z}$}
       \STATE $G_z \leftarrow \textsc{IGraph}_{\mathcal{V}}(I, z)$ \COMMENT{anchors + relations}
     \ENDFOR
     \STATE $T \leftarrow \textsc{Topo}_{\mathcal{V}}(\mathcal{Z})$
     \STATE $S \leftarrow (\mathcal{Z}, \{G_z\}_{z\in\mathcal{Z}}, T)$
     \STATE \COMMENT{\textbf{S3: Multi-Granular Intent + Derivation}}
     \FOR{$g \in \{\textsc{Coarse, Medium, Fine}\}$}
       \STATE $X \leftarrow \textsc{Prompt}_{\mathcal{V}}(I, S, g)$
       \STATE $R \leftarrow \textsc{Derive}_{\mathcal{V}}(X, S)$
       \STATE $\mathcal{D}_{ZS} \leftarrow \mathcal{D}_{ZS} \cup \{(X, S, R)\}$
     \ENDFOR
   \ENDFOR
   \STATE {\bfseries return} $\mathcal{D}_{ZS}$
\end{algorithmic}
\end{algorithm}

To address the high variance in user information density, we stratify instructions into a Granularity $\times$ Style matrix:
\begin{itemize}
  \item \textbf{Coarse (Atmospheric):} Describes only overall mood (e.g., ``A cozy vibe'') to train Design Inference for hallucinating necessary assets.
  \item \textbf{Medium (Categorical):} Identifies major furniture types to target Layout Composition.
  \item \textbf{Fine (Metric):} Imposes strict dimensional and spatial constraints to enforce Geometric Grounding.
\end{itemize}
We further inject Stylistic Diversity by randomizing syntactic openings (e.g., Imperative vs. Aspirational) and conditioning on specific aesthetic styles (e.g., Industrial $\rightarrow$ metal textures).

\subsection{Dataset Statistics}
\label{sec:appendix_data}

\begin{table}[h]
\centering
\caption{Zone-Scene-10K dataset statistics by room type.}
\begin{tabular}{lccc}
\toprule
\textbf{Room Type} & \textbf{Train} & \textbf{Val} & \textbf{Test} \\
\midrule
Bedroom & 2,845 & 165 & 300 \\
Living Room & 2,512 & 146 & 280 \\
Kitchen & 1,256 & 73 & 140 \\
Dining Room & 1,024 & 60 & 115 \\
Study/Office & 892 & 52 & 100 \\
Bathroom & 512 & 30 & 45 \\
Other & 377 & 22 & 20 \\
\midrule
\textbf{Total} & \textbf{9,418} & \textbf{548} & \textbf{1,000} \\
\bottomrule
\end{tabular}
\end{table}

\section{SCALE Benchmark Construction Details}
\label{sec:appendix_scale}

\begin{figure}[t]
\centering
\includegraphics[width=\textwidth]{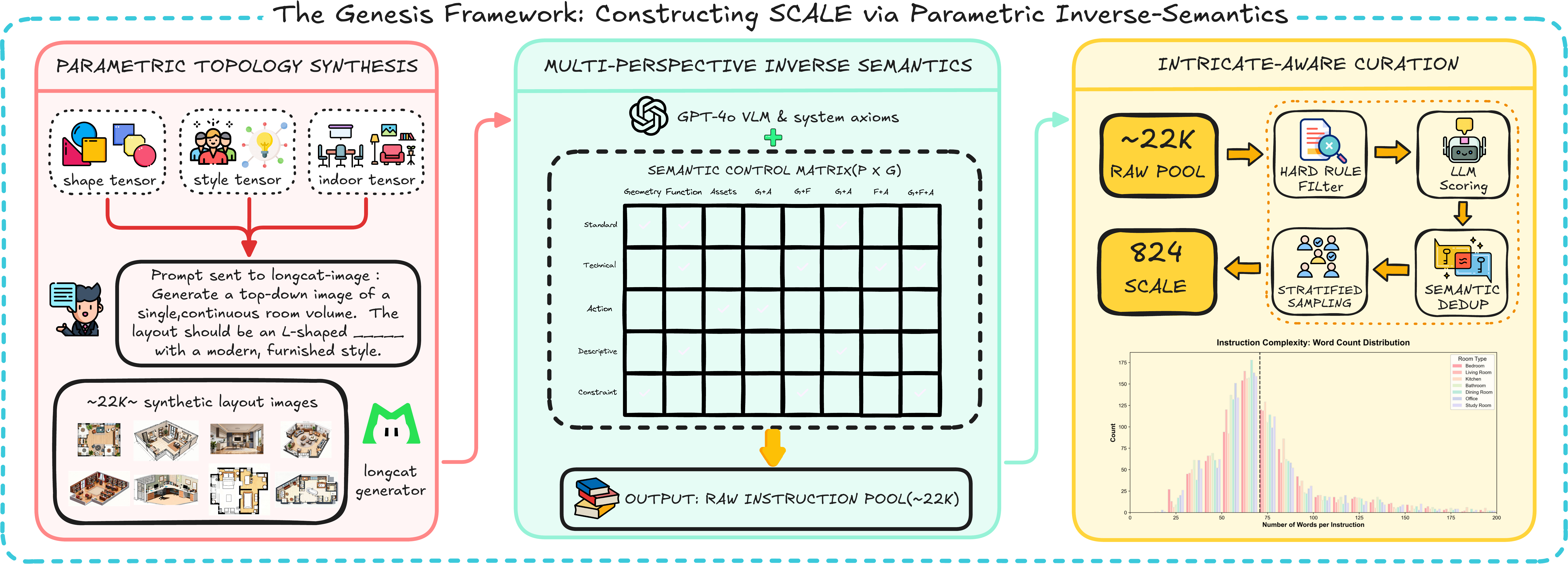}
\caption{\textbf{The SCALE Benchmark Construction Pipeline.} We employ a parametric topology sampler to generate diverse non-convex boundaries, followed by a persona-driven inverse semantics engine to synthesize multi-granular instructions.}
\label{fig:pipeline}
\end{figure}

This appendix provides comprehensive details of the SCALE Benchmark construction, including all prompts used in image generation and instruction generation, as well as the complete data cleaning pipeline. We explicitly verify that the LongCat images and prompts used to construct SCALE are strictly disjoint from the InternScenes training set to prevent data leakage.

\subsection{Floor Plan Image Generation}

\paragraph{Image Generation Model}
We use \textbf{LongCat-Image}~\cite{longcat2025image}, a 6B parameter text-to-image model from Meituan, optimized for bilingual (Chinese-English) text rendering, high photorealism output, and efficient inference ($\sim$12GB VRAM with CPU offload).

\paragraph{Generation Configuration}
\textbf{Room Shapes (9 types):}
1. Rectangular; 2. L-shaped; 3. T-shaped; 4. U-shaped; 5. H-shaped; 6. Trapezoidal; 7. Room with a diagonal wall cut; 8. Room with a protruding nook/alcove; 9. Other irregular shapes.

\textbf{Room Types (7 categories):}
Bedroom, Living Room, Kitchen, Bathroom, Dining Room, Office, Study Room.

\textbf{Interior Styles (7 variations):}
\begin{enumerate}
    \item Modern interior design, fully furnished with complete amenities
    \item Comfortable lived-in atmosphere, well-organized layout
    \item Functional layout with distinct activity zones and ample storage
    \item Spacious arrangement with multiple furniture groupings
    \item Contemporary style with detailed decor and accessories
    \item High-efficiency layout maximizing floor space utility
    \item Luxurious design with distinct separation of functions
\end{enumerate}

\textbf{Repetitions:} 5 different seeds per combination. Total Images generated: $9 \times 7 \times 7 \times 10 \times 5 = 22,050$.

\paragraph{Image Generation Prompts}
\textbf{Prefix (Shared across all images):}
\begin{quote}
``Generate a single high-quality room architectural 2D floor plan image, top-down vertical view, bird's eye view, orthographic projection, clean lines, flat shading, clearly defined walls, boundaries and furniture. The image depicts a single continuous open space defined by the outer perimeter walls only. There are absolutely no internal walls, no partitions, and no secondary rooms inside the boundary.''
\end{quote}

\textbf{Suffix Templates:}
Each suffix template contains placeholders \texttt{\{shape\}} and \texttt{\{style\}} that are dynamically filled. We utilized distinct templates for each room type to ensure diversity.

\textit{Bedroom Templates:}
\begin{enumerate}
    \item ``A \{shape\} bedroom layout. The sleeping zone is centered, with freestanding wardrobe units lining one wall. \{style\}.''
    \item ``Plan of a \{shape\} single bedroom. A study desk is positioned near the window, sharing the open space with the bed. \{style\}.''
    \item ``Top-down view of a \{shape\} bedroom. The room features a dressing area defined simply by a mirror and open clothing racks, not walls. \{style\}.''
    \item ``A \{shape\} bedroom designed for two people. Twin beds are arranged symmetrically in the single open space. \{style\}.''
    \item ``Layout of a \{shape\} bedroom where a lounge chair creates a reading nook in the corner of the room. \{style\}.''
    \item ``A large \{shape\} master bedroom. A sofa sits at the foot of the bed, creating a sitting zone within the open floor plan. \{style\}.''
    \item ``View of a \{shape\} bedroom with extensive storage cabinets arranged along the perimeter walls. \{style\}.''
    \item ``A \{shape\} bedroom with an asymmetric furniture arrangement to fit the irregular wall geometry. \{style\}.''
    \item ``A compact \{shape\} bedroom layout where the bed is tucked into a niche of the outer wall. \{style\}.''
    \item ``A \{shape\} bedroom featuring a makeup station and dresser integrated into the main sleeping area. \{style\}.''
\end{enumerate}
*(Similar templates were used for other room types, focusing on their specific furniture and functional zones.)*

\subsection{Reverse Instruction Generation}

\paragraph{Overview}
We employ GPT-4o-mini to generate natural user instructions by analyzing the generated floor plan images. This ``reverse engineering'' approach creates diverse, realistic prompts.

\paragraph{System Prompt for Instruction Generation}
\begin{small}
\begin{verbatim}
### IDENTITY & MISSION
You are an expert AI creating a high-quality dataset for 3D indoor scene generation.
Your goal is to **Simulate a User Instruction** based on a provided 2D floor plan image.

### CRITICAL TRUTH (The "Single Room" Axiom)
Before generating any text, analyze the image with these absolute rules:
1. **Single Volume:** The image depicts ONE continuous room (Bedroom, Kitchen, etc.).
2. **No Structural Partitions:** Internal lines are furniture (wardrobes, screens), 
   NOT walls.
3. **No Sub-Rooms:** Never describe separate rooms like "en-suite" or "pantry". 
   Everything is in the open plan.

### TASK PROTOCOL
You will be given:
1. **[TARGET PERSONA]:** A specific style of user (e.g., Casual, Technical).
2. **[TEMPLATE STARTER]:** The example phrase you can refer to under TARGET PERSONA.
3. **[CONTENT FOCUS]:** The specific aspect of the image to highlight 
   (Geometric Shape, Functional Zones, or Asset Density).

### EXECUTION STEPS
1. **Adopt the Persona:** Look at the [TEMPLATE STARTER]. If casual, use simple words. 
   If technical, use precise terms.
2. **Analyze the Focus:**
   - If Focus = **Geometry**: Describe the L-shape, T-shape, or irregular boundary.
   - If Focus = **Function**: Describe how furniture creates zones without walls.
   - If Focus = **Assets**: List specific furniture items and describe the density.
3. **Complete the Instruction:** 
   - Start exactly with the [TEMPLATE STARTER].
   - Continue the sentence naturally to describe the image.
   - Ensure the final output is a coherent, single-sentence command or request.

### OUTPUT FORMAT
Return **ONLY** the final completed instruction string. Do not add quotation marks.
\end{verbatim}
\end{small}

\paragraph{Content Focus Categories}
We define 7 content focus modes based on three fundamental aspects: \textbf{G} (Geometry), \textbf{F} (Function), and \textbf{A} (Assets).

\begin{table}[h]
\centering
\small
\begin{tabular}{llp{7cm}}
\toprule
\textbf{Focus} & \textbf{Name} & \textbf{Description} \\
\midrule
G & Geometry & Focus on room shape (L/T/H/Irregular) and boundaries. \\
F & Function & Focus on functional zones/activities without specific lists. \\
A & Assets & Focus on furniture lists, counts, and density. \\
G+F & Geo+Func & Combine geometric shape and functional zoning. \\
G+A & Geo+Assets & Combine room shape and asset details. \\
F+A & Func+Assets & Combine functional zoning and asset details. \\
G+F+A & Full Complex & Include all aspects. \\
\bottomrule
\end{tabular}
\end{table}

\paragraph{User Persona Styles}
We employ 7 persona styles to ensure linguistic diversity:
\begin{itemize}
    \item \textbf{Type 1: Standard Imperative} (e.g., ``Design a layout for a...'')
    \item \textbf{Type 2: Casual Conversational} (e.g., ``I'm looking for a design for a...'')
    \item \textbf{Type 3: Strictly Technical} (e.g., ``Generate an orthographic projection of a...'')
    \item \textbf{Type 4: Constraint-First} (e.g., ``Without using any internal structural walls...'')
    \item \textbf{Type 5: Action-Oriented} (e.g., ``Arrange a complete furniture set within a...'')
    \item \textbf{Type 6: Geometry-Conditional} (e.g., ``Given the specific boundary shape...'')
    \item \textbf{Type 7: Descriptive Vision} (e.g., ``A detailed top-down architectural view of a...'')
\end{itemize}

\subsection{Data Cleaning Pipeline}

\paragraph{Pipeline Overview}
The pipeline consists of 5 stages:
\begin{enumerate}
    \item \textbf{Raw Generation:} 22,050 images $\rightarrow$ 22,050 raw instructions.
    \item \textbf{Quality Evaluation:} GPT-4o-mini scoring with hard filters (Image Leak, Multi-Room, Template Violation, Length Checks).
    \item \textbf{Semantic Deduplication:} Greedy deduplication using \texttt{text-embedding-3-large} with a cosine similarity threshold of 0.8.
    \item \textbf{Data Augmentation:} Supplementing non-Geometry focus data from cache.
    \item \textbf{Balanced Sampling:} Removing simple Rectangles and uniformly sampling irregular shapes to ensure difficulty.
\end{enumerate}

\paragraph{GPT Quality Evaluation System Prompt}
\begin{small}
\begin{verbatim}
You are a data quality evaluator for a text-to-3D indoor layout generation benchmark.
Your task is to evaluate a user instruction (prompt) that was generated to describe 
a 2D floor plan image.

### EVALUATION CRITERIA
**Hard Filters (REJECT if ANY is true):**
1. IMAGE_LEAK: Contains phrases like "This image shows", "In the image".
2. MULTI_ROOM: Mentions separate rooms like "en-suite", "pantry".
3. TEMPLATE_VIOLATION: Does not start with the provided template_starter.
4. TOO_SHORT: Less than 50 characters.
5. TOO_LONG: More than 800 characters.
6. ROOM_MISMATCH: Describes wrong room type.

**Quality Score (1-10):**
1-3: Poor; 4-5: Below Average; 6-7: Good; 8-9: Very Good; 10: Excellent.

### OUTPUT FORMAT (JSON only)
{
  "pass_hard_filter": true/false,
  "reject_reason": "NONE" or [REASON],
  "quality_score": 1-10,
  "brief_comment": "One sentence reason"
}
\end{verbatim}
\end{small}

\subsection{Final Benchmark Statistics}
\textbf{Total Instructions:} 824.
\textbf{Composition:} 563 (68.3\%) contain explicit Geometry constraints; 261 (31.7\%) are control samples.

\begin{table}[h]
\centering
\caption{Distribution by Content Focus in SCALE.}
\begin{tabular}{lc}
\toprule
\textbf{Focus} & \textbf{Count} \\
\midrule
Geometry (G) & 235 \\
Geometry + Assets (G+A) & 149 \\
Geometry + Function (G+F) & 129 \\
Assets (A) & 104 \\
Function (F) & 82 \\
Function + Assets (F+A) & 75 \\
Full Complex (G+F+A) & 50 \\
\bottomrule
\end{tabular}
\end{table}

\begin{table}[h]
\centering
\small
\caption{Distribution by Room Shape in SCALE before and after deduplication.}
\begin{tabular}{lrrr}
\toprule
\textbf{Shape} & \textbf{Original} & \textbf{Deduped} & \textbf{Retention} \\
\midrule
H-shaped & 2450 & 384 & 15.7\% \\
Rectangular & 2450 & 324 & 13.2\% \\
L-shaped & 2450 & 151 & 6.2\% \\
Other irregular shapes & 2443 & 137 & 5.6\% \\
Room with a diagonal wall cut & 2450 & 92 & 3.8\% \\
T-shaped & 2450 & 75 & 3.1\% \\
Trapezoidal & 2450 & 67 & 2.7\% \\
Room with a protruding nook/alcove & 2450 & 57 & 2.3\% \\
U-shaped & 2450 & 37 & 1.5\% \\
\bottomrule
\end{tabular}
\end{table}

\section{Implementation Details}
\label{sec:appendix_impl_prompts}

This appendix provides implementation details of the ZoneMaestro framework, including the Zone-Graph annotation schema and training hyperparameters.

\subsection{Zone-Graph Annotation Schema}
\label{sec:appendix_impl}

We provide the detailed JSON schema used for Zone-Graph annotations in Zone-Scene-10K. Each scene is annotated with:
\begin{itemize}
    \item \textbf{Zone Definitions:} A list of functional zones, each containing zone ID, zone type (e.g., ``Sleeping'', ``Working'', ``Dining''), and a list of asset IDs belonging to that zone.
    \item \textbf{Intra-Zone Graph:} For each zone, a directed graph where nodes are assets and edges encode spatial relations (e.g., ``left\_of'', ``in\_front\_of'', ``on\_top\_of'').
    \item \textbf{Inter-Zone Topology:} A graph connecting zone centroids with adjacency relations (e.g., ``north\_of'', ``adjacent\_to'').
    \item \textbf{Design Monologue:} A natural language narrative explaining the design rationale, generated via reverse-engineering from the ground truth layout.
\end{itemize}

\subsection{The Group-Relative Objective Mechanics}
Unlike PPO which requires a separate Value Model (introducing training instability), GRPO estimates the baseline directly from the group mean of sampled outputs, making it highly efficient for our massive-asset generation task. For each instruction $\mathcal{X}$, we sample a group of $G$ outputs $\{Y_1, \ldots, Y_G\}$ from the reference policy $\pi_{\theta_{\text{old}}}$. The optimization objective is formulated to push the model towards layouts that are relatively better than the group average:
\begin{equation}
\mathcal{L}_{\text{GRPO}}(\theta) = \mathbb{E}_{\mathcal{X}, \epsilon} \left[ \frac{1}{G} \sum_{i=1}^{G} \min \left( r_i(\theta) \hat{A}_i, \text{clip}(r_i(\theta), 1{-}\epsilon, 1{+}\epsilon) \hat{A}_i \right) - \beta D_{\text{KL}} \right]
\end{equation}
where $r_i(\theta) = \frac{\pi_\theta(Y_i|\mathcal{X})}{\pi_{\theta_{\text{old}}}(Y_i|\mathcal{X})}$ is the importance ratio. The advantage $\hat{A}_i$ is computed by normalizing the total reward $R_i$ within the group: 
\begin{equation}
\hat{A}_i = \frac{R_i - \text{mean}(\{R_j\}_{j=1}^G)}{\text{std}(\{R_j\}_{j=1}^G)}
\end{equation}
This mechanism allows the model to explore the geometric solution space around the semantic anchor provided by SFT, effectively ``denoising'' the layout distribution.

\subsection{Hyperparameter Settings}

\paragraph{Training Configuration.}
We employ Qwen3-8B~\cite{qwen3} as our backbone foundation model, distributed across 8 NVIDIA A100 GPUs. Our training protocol follows the proposed Alternating Alignment strategy, executing two full cycles of supervised internalization followed by reinforcement alignment. For the Zone-Graph Internalization phases, we fine-tune the model for 2 epochs per cycle with a global batch size of 8 and 8 gradient accumulation steps. Subsequently, the Geometric Alignment phases employ Z-GRPO with a training batch size of 32 for 40 optimization steps. We use the AdamW optimizer with learning rates of $1e^{-5}$ for supervised phases and $5e^{-6}$ for Z-GRPO, setting the KL divergence coefficient $\beta=0.04$.

\begin{table}[h]
\centering
\caption{Hyperparameters for ZoneMaestro training.}
\begin{tabular}{lc}
\toprule
\textbf{Parameter} & \textbf{Value} \\
\midrule
Backbone Model & Qwen3-8B~\cite{qwen3} \\
Hardware & 8$\times$ NVIDIA A100 \\
Alternating Alignment Cycles & 2 \\
SFT Epochs per Cycle & 2 \\
Global Batch Size (SFT) & 8 \\
Grad Accumulation (SFT) & 8 \\
Z-GRPO Batch Size & 32 \\
Z-GRPO Optimization Steps & 40 \\
SFT Learning Rate & 1e-5 \\
Z-GRPO Learning Rate & 5e-6 \\
GRPO Group Size $G$ & 8 \\
KL Coefficient $\beta$ & 0.04 \\
$\lambda_1$ (Boundary) & 1.0 \\
$\lambda_2$ (Zone) & 0.5 \\
$\lambda_3$ (Collision) & 2.0 \\
Self-Distillation Rounds & 3 \\
\bottomrule
\end{tabular}
\end{table}

\subsection{Complete Input-Output Example}
\label{sec:appendix_io_example}

We present a complete input-output example from ZoneMaestro to illustrate the Zone-Graph reasoning process. The model receives a user instruction and generates a structured reasoning trace (\texttt{<think>}) followed by the final Zone-Graph layout (\texttt{<answer>}).

\paragraph{User Instruction (Input).}
\begin{quote}
\textit{``Make a detailed room design for a living room that uses furniture placement to create distinct functional zones, including a central conversation and relaxation area, a clearly defined children's play and reading corner along one side, and smooth circulation paths that connect these activity zones within the single open space.''}
\end{quote}

\paragraph{Zone-Graph Reasoning Trace (Output \texttt{<think>}).}
The model first analyzes the instruction and inventories required assets with their dimensions and styles. It then plans four functional zones: (1) a central conversation area anchored by an L-shaped sectional couch with satellite seating and coffee table; (2) a children's play corner with a rocking horse and books; (3) a work/laptop zone with an office chair; and (4) a storage and decor wall with cabinets and windows. The reasoning trace explicitly considers zone topology, circulation paths, and boundary constraints before committing to coordinates.

\paragraph{Zone-Graph Layout (Output \texttt{<answer>}).}
The final output is a structured JSON containing:
\begin{itemize}
    \item \textbf{Architecture}: Room boundary polygon (L-shaped, $4.78\text{m} \times 5.41\text{m}$), wall segments with normals, and window positions.
    \item \textbf{Zone Topology}: Four zones with inter-zone relations (\texttt{adjacent\_open}, \texttt{anchored\_against} walls).
    \item \textbf{Functional Zones}: Each zone contains assets with category, role (\texttt{zone\_anchor} or \texttt{satellite}), position, rotation, size, and intra-zone spatial graph edges.
\end{itemize}

This example demonstrates how ZoneMaestro's Design Monologue explicitly reasons about zone contents $\mathcal{D}$, intra-zone graphs $\mathcal{G}$, global topology $\mathcal{T}$, and architecture $\mathcal{A}$ in a causal chain before producing the final layout.

\subsection{Analysis of Structure vs. Realism Trade-off}
\label{app:structure_realism}

As noted in \cref{sec:ablation}, ZoneMaestro achieves superior Realism (4.95) and Geometric Validity but scores slightly lower on the Structure metric compared to SFT baselines. To understand this discrepancy, we analyze the definition of the ``Structure'' metric used in our GPT-4o evaluation prompt.

\vspace{5pt}
\noindent\textbf{The Definition of Structure.}
The evaluation prompt for \textit{Structural Logic} (Category 2) explicitly penalizes layouts that appear ``scattered'' or lack rigid grouping. As shown in the rubric below, the criterion favors high-level hierarchical zoning, which SFT baselines satisfy by generating sparse, grid-aligned arrangements.

\begin{tcolorbox}[colback=gray!10, colframe=gray!60, title=GPT-4o Evaluation Rubric: Structural Logic]
\small
\textbf{Category 2: Structural Logic}

\textbf{3. Structural Orchestration (Critical)}
\begin{itemize}[leftmargin=*]
    \item \textbf{Focus:} Hierarchy \& Grouping (Handling Massive Assets).
    \item \textbf{Criteria:} specifically for scenes with \textbf{massive assets (>50 items)}, does the model organize them into logical functional groups/zones? Or are they scattered randomly/piled up?
    \item \textbf{Score (0-10):} 0 = Chaotic scattering; 10 = Clear, hierarchical zoning.
\end{itemize}

\textbf{4. Geometric Grounding (Critical)}
\begin{itemize}[leftmargin=*]
    \item \textbf{Criteria:} How well does the layout adapt to \textbf{irregular geometries}?
\end{itemize}
\end{tcolorbox}

\vspace{5pt}
\noindent\textbf{Why ZoneMaestro Scores Lower.}
SFT baselines, unconstrained by physical collision checks, often produce highly symmetric, grid-like patterns that visually maximize the ``Hierarchical Zoning'' score, despite lacking physical plausibility (Realism $\approx$ 3.9). 

In contrast, ZoneMaestro is optimized via RL to ensure \textbf{zero collisions} within complex non-convex boundaries. To accommodate high asset density ($>$50 items) without intersection, the model introduces \textbf{organic irregularities}---such as slightly rotating chairs to fit alcoves or creating asymmetric clusters. While these adjustments significantly enhance \textbf{Realism} and \textbf{Physical Validity}, they increase visual entropy, which the VLM judge partially misinterprets as a reduction in ``clear hierarchical zoning.'' Thus, the lower Structure score reflects a shift from \textit{artificial rigidity} to \textit{organic, physically-grounded complexity}.

\newpage
\section{Additional Qualitative Results and Extended Case Gallery}
\label{sec:appendix_qual}

This appendix presents additional qualitative results and an extended case gallery demonstrating ZoneMaestro's capabilities across diverse room types and complexity levels. These examples further illustrate the advantages of Zone-Graph Orchestration in handling irregular geometries, maintaining zone coherence, and achieving lived-in realism.

\subsection{Additional Qualitative Comparisons}
\label{sec:appendix_qual_comparisons}
\vspace{-5pt}

\begin{figure}[H]
\centering
\vspace{-10pt}
\includegraphics[width=0.95\textwidth]{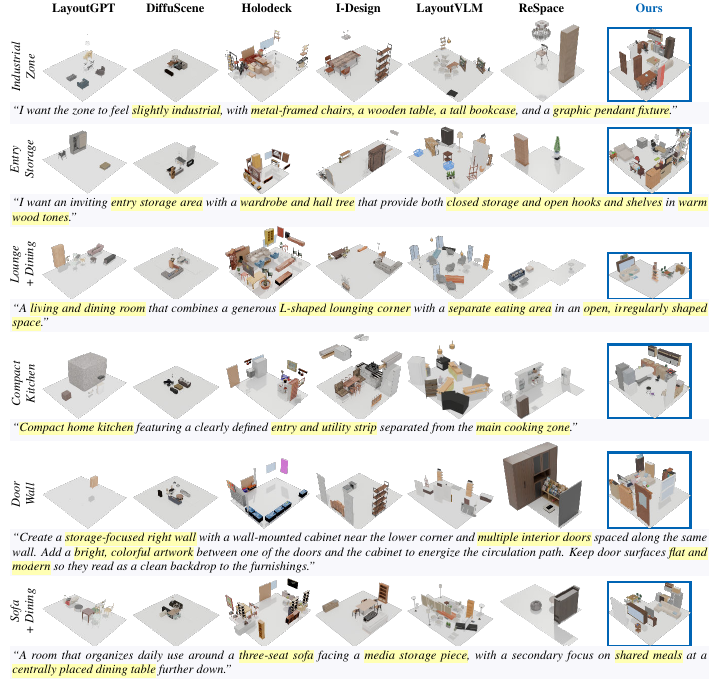}
\vspace{-5pt}
\caption{Additional qualitative comparisons with baselines on the Zone-Scene-10K Test Set (not shown in \cref{fig:qualitative}). Full user instructions are provided without abbreviation.}
\label{fig:test_appendix_cases}
\end{figure}

\vspace{-10pt}
\begin{figure}[H]
\centering
\vspace{-5pt}
\includegraphics[width=0.95\textwidth]{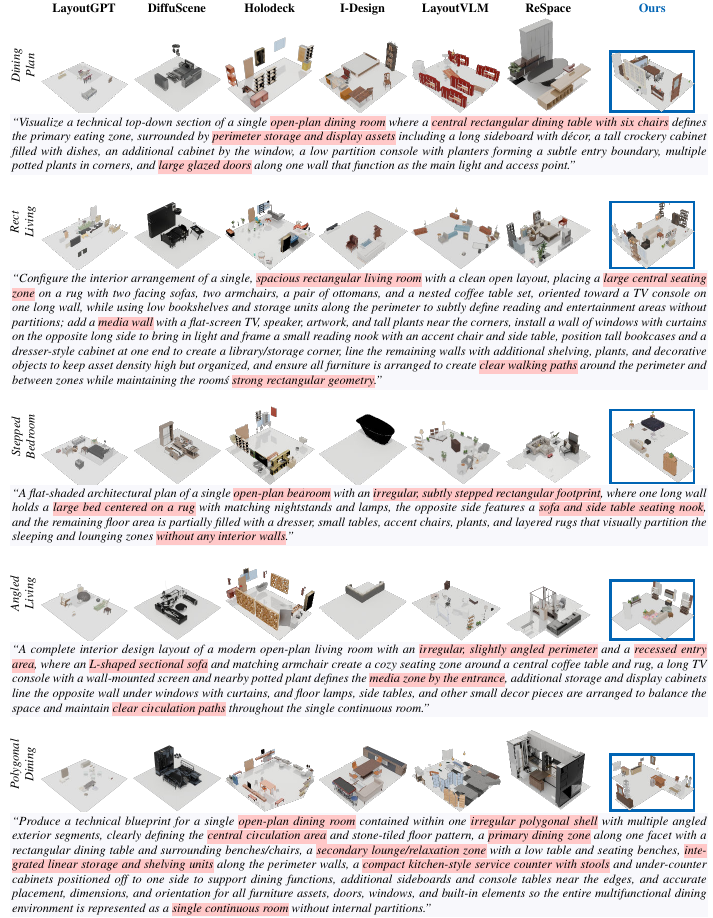}
\vspace{-5pt}
\caption{Additional qualitative comparisons with baselines on the SCALE Benchmark (not shown in \cref{fig:qualitative}). Full user instructions are provided without abbreviation.}
\label{fig:bench_appendix_cases}
\end{figure}

\subsection{Full Instructions for Main Paper Figures}
\label{sec:appendix_full_instructions}

For reproducibility, we provide the complete, unabridged user instructions corresponding to the qualitative examples shown in the main paper. \cref{fig:qualitative} includes both the Zone-Scene-10K Test Set (Rows 1--2) and the SCALE Benchmark (Rows 3--5).

\paragraph{Zone-Scene-10K Test Set (Rows 1--2)}

\textbf{Row 1 -- Multiuse Living:}
\begin{quote}
\textit{``Aiming for a multiuse living space that smoothly combines kitchen cabinets and appliances, dining and work tables, couches, stools, bins, and small decor pieces into one open room.''}
\end{quote}

\textbf{Row 2 -- L-shaped Space:}
\begin{quote}
\textit{``Create an open-plan room in an irregular L-shaped space, with a defined lounging zone and a separate area for shared meals.''}
\end{quote}

\paragraph{SCALE Benchmark (Rows 3--5)}


\textbf{Row 3 -- Open-plan Office:}
\begin{quote}
\textit{``Can you help me arrange furniture in a single open-plan rectangular office like this, with four main workstation zones along the walls (each having a long wooden desk, rolling office chair, computer monitor, keyboard, mouse, desk lamp, plants, stationery pots, and small side drawers), an L-shaped corner workstation with overhead shelves, binders, books, lamps, and a small printer cabinet, plus wall-mounted pinboards and notes above each desk to define working areas, wide glass sliding doors on two adjacent walls for natural light, a rug in the center to mark a shared circulation/collaboration space, and enough open floor area in the middle for easy movement between all the workstations without adding any interior partitions?''}
\end{quote}

\textbf{Row 4 -- Polygonal Bedroom:}
\begin{quote}
\textit{``Arrange a complete furniture set within a large irregular polygonal bedroom that widens toward the front with a long glass wall, angled rear corners, and a slightly tapered side, keeping all pieces aligned to the skewed outer perimeter.''}
\end{quote}

\textbf{Row 5 -- Pentagonal Living:}
\begin{quote}
\textit{``A high-quality 2D rendering showing a pentagonal living room with angled window walls and a mix of wood and stone flooring, where a central seating cluster of sofas, armchairs, and coffee table sits on a large rug, flanked by a long sideboard, numerous potted plants along the perimeter, and smaller accent tables that collectively fill and emphasize the unique faceted geometry of the space.''}
\end{quote}



\newcommand{\testsetheader}{%
\subsection{Extended Case Gallery}%
\label{sec:appendix_case_gallery}%
The following pages present extended qualitative results from our method.\par\vspace{0.5em}%
\noindent\textbf{Zone-Scene-10K Test Set: Extended Cases.}%
\label{sec:appendix_testset_cases}
We present additional cases from the Zone-Scene-10K test set below.}

\newcommand{\benchheader}{%
\noindent\textbf{SCALE Benchmark: Extended Cases.}%
\label{sec:appendix_bench_cases}
We present additional cases from the SCALE benchmark below.}

\newpage
\includepdf[pages=1,scale=0.75,offset=0 -2cm,pagecommand={\testsetheader}]{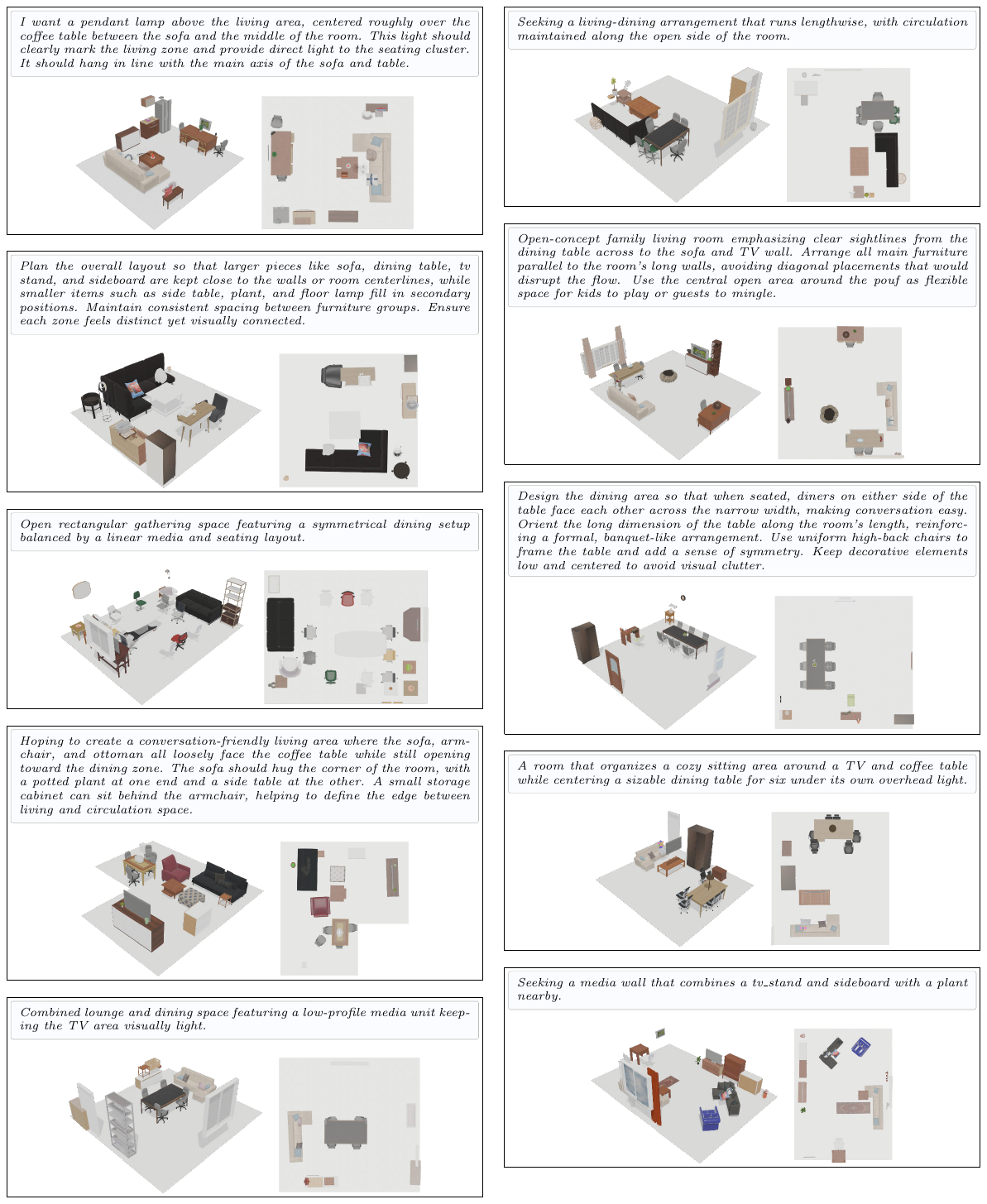}
\includepdf[pages=2-,scale=0.85,offset=0 0,pagecommand={}]{appendix_case/case_testset.pdf}

\newpage
\includepdf[pages=1,scale=0.78,offset=0 -1.5cm,pagecommand={\benchheader}]{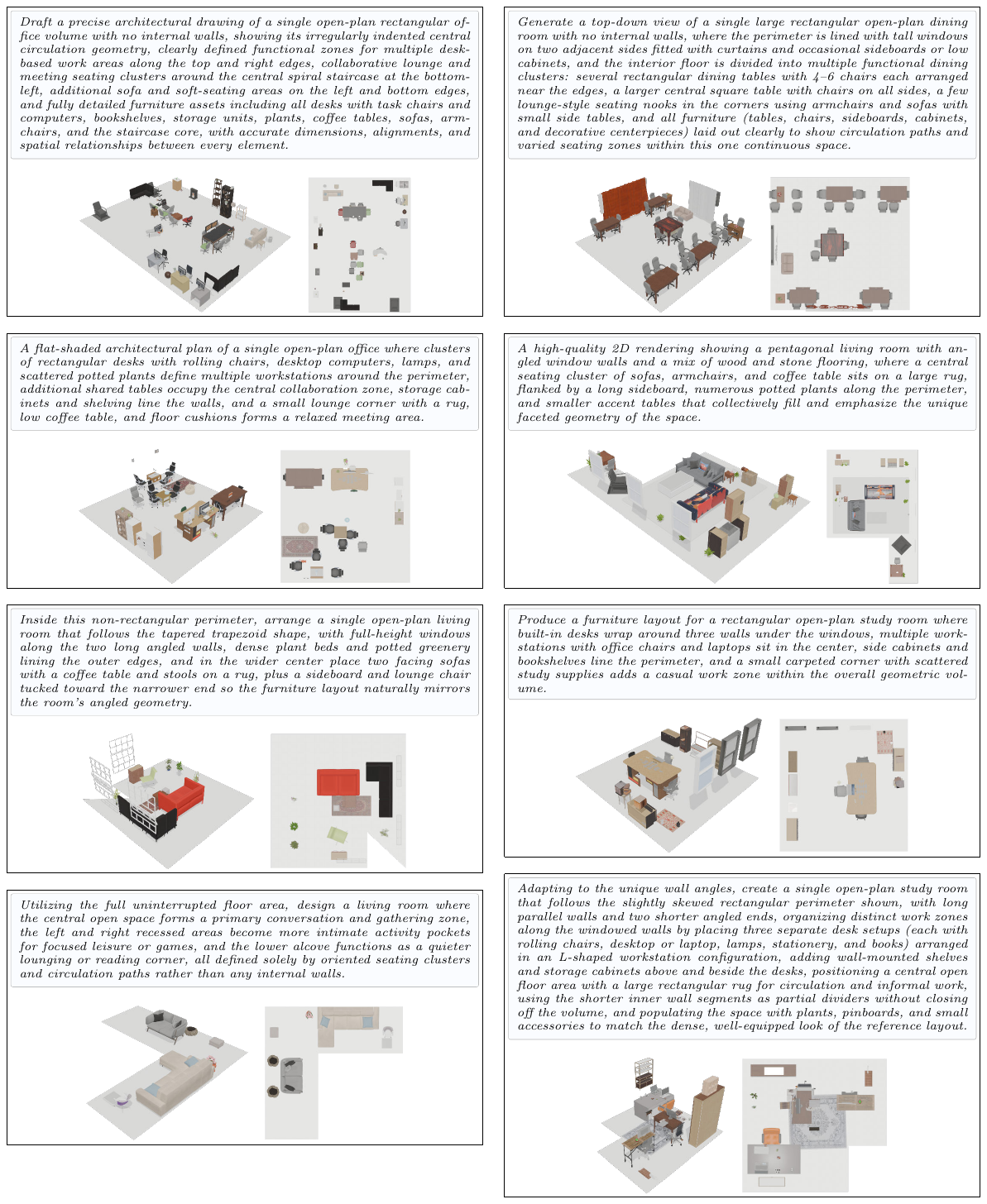}
\includepdf[pages=2-,scale=0.85,offset=0 0,pagecommand={}]{appendix_case/case_bench.pdf}


\newpage
\section{Complete Prompt Collection}
\label{sec:appendix_prompts}

This appendix provides the complete, unabridged prompts used throughout the ZoneMaestro framework. We organize them according to their role in the pipeline: Zone-Graph Annotation (\cref{sec:prompt_zonegraph}), Design Intent Synthesis (\cref{sec:prompt_userinput}), Reasoning Monologue Generation (\cref{sec:prompt_cot}), Training and Inference System Prompts (\cref{sec:prompt_system}), and Evaluation (\cref{sec:prompt_eval}).

\subsection{Zone-Graph Annotation Prompts}
\label{sec:prompt_zonegraph}

These prompts correspond to the Visual-Semantic Decomposition Pipeline described in Section~3.2. The complete pipeline operates in three stages: (1) \textbf{Visual Grouping} clusters geometrically proximate assets into candidate functional zones; (2) \textbf{Intra-Zone Spatial Graph Extraction} isolates each zone and analyzes internal spatial relationships; (3) \textbf{Global Zone Topology Derivation} establishes inter-zone adjacency and architectural anchoring relations. We present the prompts for each stage below.

\paragraph{Stage 1: Visual Grouping Prompt}
\label{sec:prompt_visual_grouping}
This prompt is used in the initial stage to cluster geometrically proximate assets into candidate functional zones based on multi-view renderings (perspective and top-down views). The output provides coarse zone boundaries that are subsequently refined.

\begin{small}
\begin{verbatim}
## Role and Goal
You are an expert AI Interior Designer and Scene Analyst. Your mission is to 
interpret a 3D indoor scene layout and reorganize its contents into functional 
groups. You will use two inputs: (1) a structured JSON file containing precise 
geometric and semantic data for each object, and (2) rendered images (a 
perspective/diagonal view and a top-down orthographic view) that provide spatial 
and stylistic context. The final output must be one valid JSON object that 
groups the original objects without altering any object data.

## Input Data Context
- Structured Layout Data (JSON): This JSON provides the ground truth for the 
  scene. Pay close attention to the `desc` (description), `pos` (position), 
  `size` (dimensions), and `jid` (unique ID) for each object:
```json
<<LAYOUT_JSON>>
```
- Visual context (rendered images): use the perspective view for holistic 
  reading and the top-down view to verify bounding-box proximity, alignment, 
  and zone boundaries.

## Core Task: Grouping Objects with Intelligence
Transform the flat objects array into a groups array reflecting human-intuitive 
functional zones by synthesizing evidence from both JSON and images.

## Guiding Principles for Grouping
- Identify functional zones (e.g., seating, dining, workspace, sleeping, 
  storage, media, decor) suggested by semantics and spatial distribution.
- Evaluate proximity using 3D bounding boxes: pos is box center; size is 
  width, height, depth; consider rotation when judging adjacency, wall-flush 
  alignment, or symmetry.
- Aim to minimize intra-group distance and maximize inter-group separation; 
  clear walkways and door swing paths often indicate boundaries.
- Canonical anchor-satellite patterns help: bed + nightstands, dining table + 
  surrounding chairs, sofa + coffee table, desk + chair, TV + media console.
- Rugs frequently bind items into one coherent zone; treat evident rug-bound 
  sets as one group.
- Alignment, symmetry around an axis/edge/centerline, facing and focal 
  relationships strengthen grouping when appropriate.

## Special Rule: Ceiling-Mounted Luminaires
Every ceiling-mounted lighting fixture (pendant, chandelier, flush or 
semi-flush ceiling light, track cluster/rail head) MUST be placed in its own 
independent group that contains only that lighting object. Do not merge 
overhead lighting into furniture-based groups. If a multi-head fixture appears 
as a single object, it still forms a single independent lighting group.

## Flexibility
It is acceptable to output a single furniture group if the space is compact 
and coherent, but overhead lighting groups must remain separate. Do not create 
empty groups.

## Inviolable Rules for Output Generation
- Perfect object integrity: the union of all group objects equals the input 
  objects exactly, one-to-one.
- No additions, deletions, or field modifications. Copy each object verbatim 
  (desc, size, pos, rot, jid).
- No duplication: an object may belong to exactly one group.

## MANDATORY PRE-OUTPUT VERIFICATION
Before finalizing your JSON output, you MUST perform this complete verification 
checklist:

1. **Object Count Verification**: Count the total number of objects in your 
   groups array. This count MUST exactly equal the number of objects in the 
   input JSON. If the counts differ, identify and fix the discrepancy.

2. **Object Completeness Check**: For EVERY object in the input JSON, verify 
   it appears exactly once in your groups array. Use the `jid` field to track 
   each object uniquely.

3. **Field Integrity Verification**: For EVERY object in your output, verify 
   that ALL fields (desc, size, pos, rot, jid) are copied character-by-character 
   identical to the input JSON. No modifications, rounding, or paraphrasing 
   allowed.

4. **No Duplication Check**: Verify that no object (identified by `jid`) 
   appears in multiple groups.

5. **No Orphaned Objects**: Ensure every object from the input appears in 
   exactly one group in your output.

If any verification step fails, you MUST correct the issue before providing 
your final JSON output.

## Additional Quality Hints
- Choose a clear anchor per group (e.g., table, bed, sofa) and gather 
  satellites via bounding-box proximity and consistent gap rules.
- Preserve functional clarity: avoid blocked access, door/drawer conflicts, 
  or overlapping use-zones between groups.
- When ambiguous, prefer the grouping with tighter internal cohesion and 
  clearer separation from neighbors.

## Output Format
Return ONLY a single valid JSON object. Do not include any text before or 
after the JSON block.

**CRITICAL: Before outputting, complete the mandatory verification checklist 
above to ensure perfect object integrity.**

```json
{
    "room_type": "", // ... from the input JSON
    "room_id": "", // ... from the input JSON
    "groups": [
        {
            "group_name": "",
            "group_type": "",
            "description": "",
            "objects": [
                // ... Verbatim object data from the input JSON
            ]
        }
        // ... other groups
    ]
}
```
\end{verbatim}
\end{small}

\paragraph{Stage 2: Intra-Zone Spatial Graph Extraction Prompt}
\label{sec:prompt_intrazone}
After obtaining coarse zone groupings from Stage 1, we render each zone in isolation by masking other zones to produce noise-free, zone-specific views. For each isolated zone, we use the following prompt to identify \textbf{Anchor Objects} (e.g., Bed, Sofa, Dining Table) and derive spatial constraints for \textbf{Satellite Objects} (e.g., Nightstand \texttt{left\_of} Bed). This prompt extracts the Intra-Zone Spatial Graph ($\mathcal{G}$) as defined in Section~3.1.

\begin{small}
\begin{verbatim}
## Role and Goal
You are an expert AI Interior Designer analyzing a SINGLE ISOLATED FUNCTIONAL 
ZONE within a larger indoor scene. Your mission is to construct the Intra-Zone 
Spatial Graph by identifying the anchor-satellite structure and deriving 
precise spatial relations between objects within this zone.

## Input Data Context
1. **Zone-Specific Layout Data (JSON)**: Contains only the objects belonging 
   to THIS zone, with `desc`, `pos`, `size`, `rot`, `model_uid` for each.
   ```json
   <<ZONE_LAYOUT_JSON>>
   ```
2. **Zone-Isolated Rendering**: A masked view showing ONLY this zone's objects,
   with other zones removed for clarity.

## Core Task: Intra-Zone Spatial Graph Construction
Analyze the isolated zone and construct a spatial graph capturing:
1. **Anchor Identification**: The primary defining object (e.g., Bed, Desk)
2. **Satellite Relations**: How secondary objects relate to the anchor
3. **Internal Spatial Constraints**: Precise geometric relationships

## Spatial Relation Taxonomy (Select Most Specific)

**Support & Containment:**
- `supported_by`: Object A rests on Object B (e.g., Lamp on Nightstand)
- `embedded_in`: Object A inside storage of B (e.g., Books in Shelf)
- `on_top_of`: Generic vertical stacking (e.g., Pillow on Bed)
- `under`: Object A underneath Object B (e.g., Rug under Table)

**Orientation & Interaction:**
- `facing_direct`: Front vector points at target (within +/-15 deg)
- `facing_angled`: Front points at target at an angle
- `back_to`: Back vector points at target
- `side_by_side`: Laterally aligned with similar orientation
- `perpendicular`: Arranged at 90 degree angle

**Arrangement Patterns:**
- `surrounding_radial`: Satellites radially around anchor (Round Table)
- `surrounding_linear`: Satellites in lines around anchor (Rect. Table)
- `flanking`: Two objects symmetrically on either side (Nightstands)

**Structure Interaction:**
- `aligned_flush`: Object back touches wall (< 5cm gap)
- `parallel_offset`: Parallel to wall with gap
- `corner_placement`: Tucked into wall corner

## Output Format
```json
{
  "zone_id": "zone_sleeping",
  "semantic_label": "Sleeping Area",
  "anchor": {
    "id": "obj_bed",
    "category": "bed",
    "description": "...",
    "transform": { "pos": [...], "rot": [...], "size": [...] }
  },
  "satellites": [
    {
      "id": "obj_nightstand_L",
      "category": "nightstand",
      "role": "satellite",
      "description": "...",
      "transform": { ... }
    }
  ],
  "spatial_graph": [
    { "source": "obj_nightstand_L", "target": "obj_bed", "relation": "flanking" },
    { "source": "obj_bed", "target": "wall_north", "relation": "aligned_flush" }
  ]
}
```
\end{verbatim}
\end{small}

\paragraph{Stage 3: Global Zone Topology Derivation Prompt}
\label{sec:prompt_topology}
After extracting the Intra-Zone Spatial Graph for each zone independently, we perform a final global analysis to derive the \textbf{Zone Topology} ($\mathcal{T}$). This prompt operates on the full scene (all zones visible) to establish inter-zone adjacency relations and zone-to-architecture anchoring constraints.

\begin{small}
\begin{verbatim}
## Role and Goal
You are an expert AI Interior Designer performing GLOBAL TOPOLOGY ANALYSIS.
Your mission is to derive the Zone Topology graph that captures inter-zone 
relationships and zone-to-architecture anchoring, completing the Hierarchical 
Scene Graph structure defined in Section 3.1.

## Input Data Context
1. **Complete Scene Layout (JSON)**: All zones with their extracted 
   Intra-Zone Spatial Graphs from Stage 2.
   ```json
   <<FULL_SCENE_JSON_WITH_ZONES>>
   ```
2. **Global Renderings**: Full scene perspective and top-down views 
   showing ALL zones and their spatial relationships.

## Core Task: Zone Topology Graph Construction
Analyze the global scene to derive:
1. **Inter-Zone Connectivity**: How zones relate to each other spatially
2. **Zone-Architecture Anchoring**: How zones attach to structural elements

## Zone Topology Relation Taxonomy

**Connectivity Relations (Zone <-> Zone):**
- `adjacent_open`: Zones touch with no barrier; uninterrupted visual flow
- `adjacent_passageway`: Connected via hallway or circulation path
- `connected_via_door`: Separated by wall but linked by door
- `separated_visual`: Share space but distinct (flooring/furniture dividers)

**Anchoring Relations (Zone <-> Structure):**
- `anchored_against`: Zone's primary furniture flush against wall
- `corner_anchored`: Zone occupies structural corner (two walls)
- `floating_center`: Zone positioned centrally, detached from walls
- `clearance_path`: Zone positioned to preserve door walkway

**Spatial Offset Descriptors:**
- `north_of`, `south_of`, `east_of`, `west_of`
- `adjacent_to`, `across_from`, `diagonal_to`

## Architectural Reconstruction
Normalize room boundary into semantically indexed nodes:
1. **Geometric Indexing**: Start from min-X vertex, traverse clockwise
2. **Wall Naming**: Sequential IDs: `wall_01`, `wall_02`, ... `wall_N`
3. **Normal Calculation**: Inward-pointing normals in Z-up system
4. **Opening Detection**: Mark passages as `opening` or `virtual_boundary`

## Output Format
```json
{
  "architecture": {
    "boundary_polygon": [[x, y, z], ...],
    "height": ...,
    "structure_nodes": [
      { "id": "wall_01", "type": "wall", "segment": [[x1,z1],[x2,z2]], 
        "normal": [nx, ny, 0] },
      { "id": "door_01", "type": "door", "pos": [x, y, z], 
        "parent_wall": "wall_03" }
    ]
  },
  "zone_topology": {
    "nodes": [
      { "id": "zone_living", "type": "primary" },
      { "id": "zone_dining", "type": "secondary" }
    ],
    "edges": [
      { "source": "zone_living", "target": "zone_dining", 
        "relation": "adjacent_open", "spatial_offset": "north_of" },
      { "source": "zone_living", "target": "wall_01", 
        "relation": "anchored_against" },
      { "source": "zone_dining", "target": "wall_02", 
        "relation": "corner_anchored" }
    ]
  }
}
```

## Verification Checklist
1. All zones from Stage 2 appear in zone_topology.nodes
2. Every zone has at least one anchoring relation to structure
3. Adjacent zones have explicit connectivity edges
4. No topology edges reference non-existent zones or walls
```
\end{verbatim}
\end{small}

\subsection{Design Intent Synthesis Prompts}
\label{sec:prompt_userinput}

These prompts are used to reverse-engineer natural user instructions from ground-truth layouts, corresponding to the ``Synthesis of Multi-Granular Design Intents'' described in Section~3.2. We provide templates for three granularity levels.

\paragraph{Coarse Granularity: Room-Level Intent}
\begin{small}
\begin{verbatim}
You are an assistant that reverse-engineers a natural, plausible interior 
design request a typical user might write. You receive a ground-truth 3D 
room layout JSON. From this, produce 16 different design briefs describing 
what is desired, implicitly matching what already exists.

INPUT LAYOUT: [GROUND_TRUTH_LAYOUT_JSON]

GRANULARITY: Coarse (room-level intent only)
- Room type and approximate size/shape in natural terms
- No object lists or spatial relations
- No style/mood/palette words or qualifiers
- Aim for 1 sentence per brief

OPENING STYLE: [OPENING_STYLE]

**FORBIDDEN:**
- makeover/alteration verbs implying an existing space ("transform", 
  "redesign", "remodel", "convert", "reconfigure", "turn into", "make over", 
  "upgrade", "refresh")
- infinitive openings (e.g., starting with "To ...")
- Raw JSON keys, coordinates, quaternions, IDs, asset codes
- Mentions of inputs, data, layouts, JSON, or "according to"
- Meta phrases ("based on the JSON," "according to the layout")
- Long copied object descriptions
- Over-claiming unseen features (windows, views, ceiling type) unless 
  unmistakably implied

Note: **the floor of the room is always a flat plane, so no need to mention 
it.**

OUTPUT FORMAT:
- 16 numbered design briefs. Each brief is one sentence. Use the format 
  "1. [brief]", "2. [brief]", etc. No other headings or labels.
\end{verbatim}
\end{small}

\paragraph{Medium Granularity: Object Categories}

\textbf{Without Style:}
\begin{small}
\begin{verbatim}
You are an assistant that reverse-engineers a natural, plausible interior 
design request a typical user might write. You receive a ground-truth 3D 
room layout JSON. From this, produce 16 different concise design briefs 
describing what is desired, implicitly matching what already exists.

INPUT LAYOUT: [GROUND_TRUTH_LAYOUT_JSON]

GRANULARITY: Medium (object categories, no style)
- List main object categories (category-level nouns only; no sizes/counts/
  brands/models)
- No placements/relations
- NO style/mood/palette hints or feature qualifiers
- Use plain category nouns only
- Aim for 1-2 sentences per brief

OPENING STYLE: [OPENING_STYLE]

**FORBIDDEN:**
- makeover/alteration verbs implying an existing space ("transform", 
  "redesign", "remodel", "convert", "reconfigure", "turn into", "make over", 
  "upgrade", "refresh")
- infinitive openings (e.g., starting with "To ...")
- Raw JSON keys, coordinates, quaternions, IDs, asset codes
- Mentions of inputs, data, layouts, JSON, or "according to"
- Meta phrases ("based on the JSON," "according to the layout")
- Long copied object descriptions
- Over-claiming unseen features (windows, views, ceiling type) unless 
  unmistakably implied

Note: **the floor of the room is always a flat plane, so no need to mention 
it.**

OUTPUT FORMAT:
- 16 numbered design briefs. Each brief is 1-2 sentences. Use the format 
  "1. [brief]", "2. [brief]", etc. No other headings or labels.
\end{verbatim}
\end{small}

\textbf{With Style:}
\begin{small}
\begin{verbatim}
You are an assistant that reverse-engineers a natural, plausible interior 
design request a typical user might write. You receive a ground-truth 3D 
room layout JSON. From this, produce 16 different concise design briefs 
describing what is desired, implicitly matching what already exists.

INPUT LAYOUT: [GROUND_TRUTH_LAYOUT_JSON]

GRANULARITY: Medium (object categories, with style)
- List main object categories (category-level nouns only; no sizes/counts/
  brands/models)
- No placements/relations
- May include short style/mood/palette hint
- Keep style minimal and abstract
- Aim for 1-2 sentences per brief

OPENING STYLE: [OPENING_STYLE]

**FORBIDDEN:**
- makeover/alteration verbs implying an existing space ("transform", 
  "redesign", "remodel", "convert", "reconfigure", "turn into", "make over", 
  "upgrade", "refresh")
- infinitive openings (e.g., starting with "To ...")
- Raw JSON keys, coordinates, quaternions, IDs, asset codes
- Mentions of inputs, data, layouts, JSON, or "according to"
- Meta phrases ("based on the JSON," "according to the layout")
- Long copied object descriptions
- Over-claiming unseen features (windows, views, ceiling type) unless 
  unmistakably implied

Note: **the floor of the room is always a flat plane, so no need to mention 
it.**

OUTPUT FORMAT:
- 16 numbered design briefs. Each brief is 1-2 sentences. Use the format 
  "1. [brief]", "2. [brief]", etc. No other headings or labels.
\end{verbatim}
\end{small}

\paragraph{Fine Granularity: Objects and Spatial Relations}

\textbf{Without Style:}
\begin{small}
\begin{verbatim}
You are an assistant that reverse-engineers a natural, plausible interior 
design request a typical user might write. You receive a ground-truth 3D 
room layout JSON. From this, produce 16 different concise design briefs 
describing what is desired, implicitly matching what already exists.

INPUT LAYOUT: [GROUND_TRUTH_LAYOUT_JSON]

GRANULARITY: Fine (objects + spatial relations, no style)
- Category-level nouns with short relative placements between major 
  objects/groups/zones
- No coordinates/angles/numeric dimensions; keep relations high-level 
  and plausible
- NO style/mood/palette hints or feature qualifiers
- Use plain category nouns only
- Aim for 2-5 sentences per brief

OPENING STYLE: [OPENING_STYLE]

**FORBIDDEN:**
- makeover/alteration verbs implying an existing space ("transform", 
  "redesign", "remodel", "convert", "reconfigure", "turn into", "make over", 
  "upgrade", "refresh")
- infinitive openings (e.g., starting with "To ...")
- Raw JSON keys, coordinates, quaternions, IDs, asset codes
- Mentions of inputs, data, layouts, JSON, or "according to"
- Meta phrases ("based on the JSON," "according to the layout")
- Long copied object descriptions
- Over-claiming unseen features (windows, views, ceiling type) unless 
  unmistakably implied

Note: **the floor of the room is always a flat plane, so no need to mention 
it.**

OUTPUT FORMAT:
- 16 numbered design briefs. Each brief is 2-5 sentences. Use the format 
  "1. [brief]", "2. [brief]", etc. No other headings or labels.
\end{verbatim}
\end{small}

\textbf{With Style:}
\begin{small}
\begin{verbatim}
You are an assistant that reverse-engineers a natural, plausible interior 
design request a typical user might write. You receive a ground-truth 3D 
room layout JSON. From this, produce 16 different concise design briefs 
describing what is desired, implicitly matching what already exists.

INPUT LAYOUT: [GROUND_TRUTH_LAYOUT_JSON]

GRANULARITY: Fine (objects + spatial relations, with style)
- Category-level nouns with short relative placements between major 
  objects/groups/zones
- No coordinates/angles/numeric dimensions; keep relations high-level 
  and plausible
- May include short style/mood/palette hint
- Keep style minimal and abstract
- Aim for 2-5 sentences per brief

OPENING STYLE: [OPENING_STYLE]

**FORBIDDEN:**
- makeover/alteration verbs implying an existing space ("transform", 
  "redesign", "remodel", "convert", "reconfigure", "turn into", "make over", 
  "upgrade", "refresh")
- infinitive openings (e.g., starting with "To ...")
- Raw JSON keys, coordinates, quaternions, IDs, asset codes
- Mentions of inputs, data, layouts, JSON, or "according to"
- Meta phrases ("based on the JSON," "according to the layout")
- Long copied object descriptions
- Over-claiming unseen features (windows, views, ceiling type) unless 
  unmistakably implied

Note: **the floor of the room is always a flat plane, so no need to mention 
it.**

OUTPUT FORMAT:
- 16 numbered design briefs. Each brief is 2-5 sentences. Use the format 
  "1. [brief]", "2. [brief]", etc. No other headings or labels.
\end{verbatim}
\end{small}

\subsection{Reasoning Monologue Generation Prompt}
\label{sec:prompt_cot}

This prompt generates the Design Monologue ($\mathcal{R}$) that serves as the reasoning trace for Zone Reasoning Internalization (Section~3.3). It reverse-engineers a coherent design narrative from the ground-truth layout.

\begin{small}
\begin{verbatim}
You are an expert 3D indoor scene layout designer using a Structured 
Hierarchical Spatial Reasoning (SHSR) process. Generate a single, cohesive, 
plain text monologue that reads like a designer's internal narrative. It 
must begin from a user brief and naturally arrive at the exact final layout, 
without revealing that any ground truth exists.

For your internal use only (never mention explicitly in the output):
Design Brief: <<<DESIGN_BRIEF_HERE>>>
Target Layout (Hidden Ground Truth): <<<TARGET_LAYOUT_JSON_HERE>>>
Target Renders (Hidden Ground Truth): diagonal and top views of the final 
scene

Primary objective:
Generate a design monologue that simulates a complete reasoning process. 
The monologue must start by interpreting the design brief, inventorying all 
objects with their final attributes, reasoning through functional zones and 
per-object placement in a structured way, then establishing and validating 
the room architecture boundary and vertical room volume so that they exactly 
accommodate the final layout. This process must culminate in a final 
arrangement where every size, position, rotation, and grouping precisely 
matches the hidden target layout, with the entire thought process appearing 
self-motivated and plausible.

Hard style constraints (output must follow ALL):
Language density control: Throughout the monologue, avoid filler determiners 
such as "the", "a", "an" unless absolutely required for grammar. Favor direct 
noun phrases, varied sentence structures, and concrete references. Keep prose 
natural but tighten unnecessary fillers. Aim for clarity and flow without 
relying on constant determiners.
Plain text paragraphs only. No headings, no numbered steps, no bullet points, 
no code blocks, no JSON, no tables.
Do not use labels like "Step", "Thought", or any list markers (for example 
leading dashes or numbers). Do not use symbols that look like markdown 
headings or code fences.
Do not include raw keys, IDs, asset codes, or meta phrases that expose 
implementation such as data formats or file types.
Prefer paragraphs that feel like a real-time design process.
When you mention dimensions, positions, or rotations, weave them into 
sentences (for example "I place the sofa at pos [x, y, z] with rot 
[rx, ry, rz]"), never as lists or headings.
Final numeric values for all objects and the room architecture (boundary 
and height) must be exactly those of the hidden target.

Technical conventions to apply (do not restate as headings; just use them 
consistently):
Coordinate system: right-handed, Z-up; floor plane z=0; units in meters.
Rotations: 3D Euler angles [rx, ry, rz] in radians.
Room architecture and boundary: a vertical room volume defined by a floor 
polygon and a matching ceiling polygon; identical vertex count and one to 
one correspondence; the floor polygon lies on z=0; the ceiling polygon lies 
on z=H (H>0). This architectural boundary is a strict limit; no part of any 
object's bounding box may extend beyond it.
Treat walls and floor as abstract boundaries; avoid over specifying 
architectural details.

Reasoning process guidance (write fluid prose; do not label these as steps):
Interpret the brief in reasoning mode with one consistent pipeline and 
assume it is consistent with the target. For coarse briefs, expand high 
level intent into a functional program, zone hypotheses, adjacency rules 
and arrangement principles before any precise placement. For medium briefs, 
complete missing constraints such as zoning detail, anchor choices, axis or 
symmetry, and walkway and clearance policies that lead to final placements. 
For fine briefs, treat given specifics as binding and add only minimal 
offsets, clearances and small rotations needed to reach exact placements. 
When enriching, infer only what is necessary and keep additions neutral and 
plausible. State guiding principles such as alignment, symmetry, adjacency, 
balanced sightlines and ergonomic clearances.

Object Inventory, Attributes, and Sizing: 
Pre-inventory all objects with their exact dimensions (size_x x size_y x 
size_z), presenting this as a designer's preparatory list justified by the 
brief's requirements. For each object, also articulate concise visual and 
physical attributes a designer would use: color and finish palette, form 
factor or shape geometry, style vocabulary, texture or materiality, and any 
distinctive features or affordances such as rounded corners, tufted 
upholstery, slatted doors, tapered legs, or glass top. Keep one or two short 
phrases per attribute; be consistent with the brief; avoid IDs or raw keys; 
and conclude with a one or two sentence natural-language summary that 
synthesizes these attributes to describe the object.
Zone by zone and object by object layout reasoning: reason through functional 
zones sequentially, following a hierarchical logic similar to a scene graph. 
For each zone, choose a clear functional role and an anchor object grounded 
in the brief and room type, then reason out a plausible initial pos and rot 
for that anchor that respects ergonomic reach, wall relationships and 
sightlines. Within each zone, perform object by object placement reasoning: 
for every asset in that zone, explicitly derive and state a concrete pos 
[x, y, z] and rot [rx, ry, rz], justified by ergonomic adjacency, walkway 
and clearance policy, facing and focal direction, axis or edge or centerline 
alignment, balanced sightlines, and alignment with human interior design 
preferences. Each placement must account for all previously placed objects 
in the same zone, preserve zero collision with previously placed assets, 
respect functional separation between zones to avoid blocked access or task 
interference, and maintain generous, breathable spacing rather than crowding. 
Use bounding box proximity, consistent gap rules, stable wall offsets, and 
ergonomic ranges for circulation and reach to keep the configuration coherent. 
Do not rely on intentional numeric perturbations or temporary inconsistencies 
relative to the hidden target; keep the reasoning narrative smooth and 
convergent toward the exact final configuration.
Next, establish and verify the foundational room architecture and boundary 
after you have reasoned through functional zoning and per-object placement. 
Derive precise room height and floor and ceiling polygon vertices from the 
spatial requirements of the complete object collection, the inferred room 
type, and user intent. Justify why this exact architectural boundary tightly 
but comfortably houses all furniture and circulation, and explicitly confirm 
in the narrative that every object's bounding box remains strictly and 
entirely inside this boundary with appropriate clearances to walls and edges.

Final scene level verification: 
conduct a concluding check to confirm the layout's success. This includes 
verifying generous ergonomic clearances, absolutely zero collisions between 
any 3D objects, clear and coherent circulation paths, and strict containment 
of every object's bounding box within the room architectural boundary. The 
final monologue should confirm that the design fully and elegantly fulfills 
the initial brief while matching the intended final layout.
Functional grouping and zone delineation: articulate why the placed objects 
cohere into logical functional zones and how these zones are positioned 
relative to each other and to the architecture to create a clear, legible 
layout. Explain how anchors, satellites, adjacency, and separation follow 
consistent principles similar to a hierarchical scene graph, and how the 
final arrangement respects room boundaries and supports human activities 
implied by the brief.

Self-check before finalizing:
Ensure no line in the monologue starts with symbols or patterns that would 
be interpreted as headings, list markers, or code fences. Ensure you have 
explicitly given sizes for all objects and final pos and rot [rx, ry, rz] 
for all adjustable placements embedded naturally in sentences. Do not 
mention any provided data, file formats, or images; keep the entire answer 
as one plain text monologue.
\end{verbatim}
\end{small}

\subsection{Training and Inference System Prompt}
\label{sec:prompt_system}

This is the system prompt used during both supervised fine-tuning (SFT), reinforcement learning (Z-GRPO), and inference. It defines the output contract and reasoning protocol for the ZoneMaestro model.

\begin{small}
\begin{verbatim}
You are an expert AI Interior Architect. Your task is to generate a complete 
3D indoor scene layout from a user's design brief through structured reasoning 
and thoughtful design decisions.
 
Output contract:
- First, produce your design reasoning process enclosed in <think>...</think>.
- Then, produce the final layout as a single valid JSON object enclosed in 
  <answer>...</answer>.
 
Design reasoning process for <think>:
- Think and reason through the design challenge systematically, demonstrating 
  how you arrive at each decision.
- Write as a continuous internal dialogue that shows your thought progression 
  from understanding the brief to finalizing the layout.
- Your reasoning should feel like a designer thinking through the problem in 
  real-time, making decisions, evaluating them, and refining as needed.
 
Technical conventions (apply consistently throughout your reasoning):
- Coordinate system: right-handed, Z-up; floor plane z = 0; units in meters.
- Rotations: Euler angles [x, y, z].
- Room architecture: a vertical prism with congruent top/bottom polygons; 
  same vertex count and one-to-one correspondence; bounds_bottom vertices lie 
  on z = 0; bounds_top on z = H (H > 0).
- Ergonomics: maintain clear walkways, avoid any object-to-object collisions 
  and object-to-boundary violations, preserve functional adjacency, 
  comfortable reach distances, and coherent sightlines.
 
Reasoning flow guidance for <think> (think through these aspects naturally):
- Think and reason with one consistent pipeline regardless of how detailed 
  the brief is. For vague briefs, expand high-level intent into a functional 
  program, zone hypotheses, adjacency rules, and arrangement principles 
  before any precise placement. For detailed briefs, identify the binding 
  constraints and think through how to operationalize them while filling in 
  missing details like exact positions and clearances.
- Define the room architecture by reasoning about the space needed to 
  accommodate all required functions with proper circulation.
- Think through your object inventory, reasoning about appropriate sizes 
  (W x H x D) and visual attributes for each item. For each object, consider 
  color/finish palette, form factor/shape geometry, style vocabulary, 
  texture/materiality, and any distinctive features or affordances. 
  Synthesize these into a natural description.
- Reason through placement decisions zones by zones. For each functional 
  zone, think about the anchor object, then systematically place related 
  items. As you place each object, explicitly state its pos [x, y, z] and 
  rot [x, y, z] and justify it against established principles. Your 
  reasoning must confirm that each new placement avoids collision with the 
  room architecture, previously placed zones, and other objects within its 
  own zone.
- After initial placement reasoning, critically evaluate your layout. When 
  you identify issues with circulation, alignment, balance, functional 
  relationships, or any collision or boundary violation, start the next 
  sentence with "wait..." and think through corrections, then restate the 
  improved pos and rot values.
- Conclude by verifying that your reasoning has led to a coherent, 
  collision-free design that fulfills the brief and respects all spatial 
  boundaries.
 
Design principles to apply in your reasoning:
- Strict architecture Containment: All assets must be fully contained within 
  the room architecture without exception.
- Zero Collisions: The layout must be free of unintended collisions. This 
  includes inter-zones (between zones), and intra-zone (within a zone) 
  collisions.
- Clear Circulation: Maintain ergonomic walkways and comfortable clearances 
  for access and movement.
- Functional Adjacency: Position related objects logically to support their 
  intended use.
- Proximity without Crowding: Zone items closely to create functional zones, 
  but maintain enough space to avoid a cluttered feel.
- Balanced Composition: Distribute visual weight to create a sense of 
  harmony and stability.
- Alignment and Symmetry: Use shared axes, edges, or centerlines to create 
  order, but only where appropriate for the design style.
- Logical Facing and Sightlines: Orient objects to support interaction 
  (e.g., conversational seating) and create pleasing views.
 
Contents for the <answer> tag: The Final JSON Layout
- Return a single, valid JSON object only (no extra text), conforming to 
  this shape:
 
{
  "meta": {
    "scene_type": "string"
  },
  "architecture": {
    "boundary_polygon": [[x, y, z], ...],
    "structure_nodes": [
      {
        "id": "...",
        "type": "...",
        "segment": [[x1, z1], [x2, z2]],
        "normal": [x, y, z]
      },
      { "id": "door_1", "type": "door", "pos": [x, y, z] }
    ]
  },
  "zone_topology": {
    "nodes": [
      { "id": "zone_1", "type": "primary" },
      { "id": "zone_2", "type": "secondary" }
    ],
    "edges": [
      {
        "source": "zone_1",
        "target": "zone_2",
        "relation": "adjacent_open",
        "spatial_offset": "north_of"
      },
      {
        "source": "zone_1",
        "target": "wall_north_main",
        "relation": "anchored_against"
      }
    ]
  },
  "functional_zones": [
    {
      "id": "...",
      "semantic_label": "...",
      "assets": [
        {
          "id": "obj_1",
          "category": "...",
          "role": "...",
          "description": "...",
          "pos": [x, y, z],
          "rot": [rx, ry, rz],
          "size": [w, h, d]
        }
      ],
      "spatial_graph": [
        {
          "source": "obj_x",
          "target": "obj_y",
          "relation": "..."
        },
        {
          "source": "obj_m",
          "target": "...",
          "relation": "..."
        }
      ]
    }
  ]
}
 
JSON validity requirements:
- The JSON must be syntactically valid (numeric fields are numbers; no 
  trailing commas; no comments).
- Top and bottom polygons must have identical vertex counts and correspond 
  1:1; all bounds_bottom vertices must have z = 0; all bounds_top vertices 
  must share the same z = H.
- Every object's axis-aligned bounding box must lie within the room 
  architecture.
- Floor-standing items should have their bottom at z = 0; wall-mounted 
  items should have appropriate heights and rotations.
- Objects should be grouped logically by function; each object appears 
  exactly once across all zones.
- The layout should respect ergonomic clearances and be free of unintended 
  collisions.
\end{verbatim}
\end{small}

\subsection{GPT-4o-mini Evaluation Prompt}
\label{sec:prompt_eval}

This prompt is used for the GPT-4o-mini judge that evaluates generated layouts across six metrics organized into three categories: Perceptual Quality, Structural Logic, and Semantic Accuracy (Section~5.1).

\begin{small}
\begin{verbatim}
# Role Definition
You are an expert Senior Architect and Spatial Planner. Your task is to 
evaluate a generated 3D indoor scene based on the provided visualization 
renderings and the user's text instruction.

# Input Data
1. **Text Instruction:** The original prompt describing the scene (e.g., 
   "A cluttered, L-shaped artist studio with over 50 items").
2. **Visual Renderings:** Perspective images of the generated scene.

# Critical Constraints (READ CAREFULLY)
- **IGNORE Rendering Quality:** Do NOT downgrade scores for low resolution, 
  blur, pixelation, or lighting artifacts.
- **IGNORE Asset Texture:** Do NOT evaluate the material quality or texture 
  resolution of the furniture.
- **FOCUS ONLY ON:** Spatial layout, geometric logic, object arrangement, 
  and instruction adherence.
- **Scoring Scale:** Provide an **INTEGER score from 0 to 10** for EACH of 
  the 6 metrics below (0 = Failure, 10 = Perfect).

# Evaluation Metrics

Please evaluate the scene across the following 3 categories and 6 specific 
metrics:

## Category 1: Perceptual Quality
**1. Aesthetic Harmony**
*   **Focus:** Visual Style & Consistency.
*   **Criteria:** Is the visual style consistent across the room? Do the 
    furniture pieces stylistically belong together? This is a baseline 
    quality check for visual coherence.
*   **Score (0-10):** 0 = Mismatched, chaotic styles; 10 = Perfectly unified 
    stylistic theme.

**2. Lived-in Realism (Critical)**
*   **Focus:** Organic Entropy vs. Synthetic Showroom.
*   **Criteria:** Does the scene look like a real, inhabited space with 
    natural "clutter" and organic variation? Or does it look like a sterile, 
    artificial AI-generated showroom with rigid, grid-like alignment?
*   **Score (0-10):** 0 = Artificial, robotic alignment, sterile; 
    10 = Highly organic, natural rotations, convincing "lived-in" vibe.

## Category 2: Structural Logic
**3. Structural Orchestration (Critical)**
*   **Focus:** Hierarchy & Grouping (Handling Massive Assets).
*   **Criteria:** specifically for scenes with **massive assets (>50 items)**, 
    does the model organize them into logical functional groups/zones? Or 
    are they scattered randomly/piled up?
*   **Score (0-10):** 0 = Chaotic scattering or overlapping piles; 
    10 = Clear, hierarchical zoning of many objects.

**4. Geometric Grounding (Critical)**
*   **Focus:** Boundary Adaptation (Non-Convex Rooms).
*   **Criteria:** How well does the layout adapt to **irregular geometries** 
    (e.g., L-shaped, H-shaped, alcoves)? Does it utilize nooks effectively, 
    or do objects float in void spaces/clip through irregular walls?
*   **Score (0-10):** 0 = Ignores room shape, severe clipping/floating; 
    10 = Perfect adaptation to the specific non-convex boundary.

## Category 3: Semantic Accuracy
**5. Semantic Fidelity**
*   **Focus:** Instruction Following.
*   **Criteria:** Does the scene strictly contain the room type and specific 
    objects requested in the text prompt?
*   **Score (0-10):** 0 = Completely wrong room/objects; 10 = Perfect recall 
    of all requested elements.

**6. Functional Affordance**
*   **Focus:** Physics & Navigation.
*   **Criteria:** Is the layout physically plausible and navigable? Are 
    paths clear? Are objects placed logically for human use (e.g., chairs 
    facing tables)?
*   **Score (0-10):** 0 = Blocked paths, physically impossible placements; 
    10 = Highly functional and navigable.

# Output Format
Provide your evaluation in the following JSON format:

```json
{
  "perceptual": {
    "aesthetic_harmony_score": <int>,
    "lived_in_realism_score": <int>,
    "reasoning": "<Brief explanation for perceptual scores>"
  },
  "structural": {
    "structural_orchestration_score": <int>,
    "geometric_grounding_score": <int>,
    "reasoning": "<Brief explanation for structural scores>"
  },
  "semantic": {
    "semantic_fidelity_score": <int>,
    "functional_affordance_score": <int>,
    "reasoning": "<Brief explanation for semantic scores>"
  }
}
```
\end{verbatim}
\end{small}

\section{Limitations}
\label{sec:appendix_limitations}

While ZoneMaestro advances intricate scene orchestration, we acknowledge limitations pointing toward future research. Our framework currently prioritizes global structural coherence and static physical validity. It does not explicitly model kinematic articulation such as the swing radius of doors or drawers, which leaves fine-grained interactive physics for downstream refinement. Additionally, the method focuses on spatial intelligence rather than texture synthesis or lighting simulation. The generated layouts serve as geometric scaffolds that require integration with separate material pipelines for photo-realistic rendering. Finally, the system relies on semantic priors distilled from residential and commercial scans. Generalization to highly abstract domains beyond typical architectural forms remains bounded by the training distribution and would benefit from domain-specific data injection.

\end{document}